\documentclass{article}
\usepackage[T1]{fontenc}
\usepackage{iclr2027_conference,times}

\usepackage{amsmath,amssymb}
\usepackage{graphicx}
\usepackage{booktabs}
\usepackage{xcolor}
\usepackage{placeins}
\usepackage{capt-of}
\usepackage{wrapfig}

\usepackage{tikz}
\newlength{\ablcellw}
\newlength{\ablcellh}
\newcommand{\nacell}{%
  \setlength{\ablcellw}{0.155\linewidth}%
  \setlength{\ablcellh}{0.6657\ablcellw}%
  \begin{tikzpicture}
    \draw[gray!60] (0,0) rectangle (\ablcellw,\ablcellh);
    \draw[gray!60] (0,0) -- (\ablcellw,\ablcellh);
    \draw[gray!60] (0,\ablcellh) -- (\ablcellw,0);
  \end{tikzpicture}%
}
\usepackage{url}
\definecolor{citeblue}{rgb}{0.0,0.30,0.70}
\usepackage[colorlinks=true,citecolor=citeblue,linkcolor=black,urlcolor=black]{hyperref}

\usepackage{titletoc}
\newcommand{\tocleader}{{\normalfont\titlerule*[0.75em]{.}}}
\titlecontents{section}[2.0em]
  {\addvspace{7pt}\normalsize\bfseries}
  {\contentslabel{2.0em}}
  {\hspace*{-2.0em}}
  {\tocleader\contentspage}
  [\addvspace{1pt}]
\titlecontents{subsection}[4.6em]
  {\addvspace{1pt}\normalsize}
  {\contentslabel{2.6em}}
  {\hspace*{-2.6em}}
  {\tocleader\contentspage}

\usepackage{letltxmacro}
\newcommand{\refbox}[1]{{\setlength{\fboxsep}{0.6pt}\setlength{\fboxrule}{0.5pt}%
  \fcolorbox{red}{white}{#1}}}
\makeatletter
\AtBeginDocument{%
  \LetLtxMacro\LLorigref\ref
  \DeclareRobustCommand{\ref}{\@ifstar{\LLorigref*}{\LL@boxedref}}%
}
\newcommand{\LL@boxedref}[1]{\refbox{\LLorigref{#1}}}
\makeatother

\newcommand{\trilerp}{\operatorname{trilerp}}
\newcommand{\supp}{\operatorname{supp}}
\newcommand{\Ilo}{\mathbf{I}_{\downarrow}}

\title{LoopLUT: 3D Lookup Tables with\\
  Progressive Region Refinement for\\
  Real-Time 4K Image Enhancement}

\author{Yang Ye$^{1*}$, Jiajun Ma$^{1*}$, Chen Wu$^{2}$, Wei Wang$^{3}$, Dianjie Lu$^{4}$, Guijuan Zhang$^{4}$, \\
\bf Linwei Fan$^{5}$, Zhuoran Zheng$^{2\dagger}$ \\
{\small $^{1}$Universiti Sains Malaysia \quad $^{2}$National University of Defense Technology \quad $^{3}$Sun Yat-sen University} \\
{\small $^{4}$Shandong Normal University \quad $^{5}$Shandong University of Finance and Economics}}

\iclrfinalcopy

\begin{document}

\maketitle
\lhead{}
\renewcommand{\headrulewidth}{0pt}
{\renewcommand{\thefootnote}{}\footnotetext{$^{*}$Equal contribution. \quad $^{\dagger}$Corresponding author.}}

\begin{abstract}
Color enhancement of 4K images must meet a quality target under a tight compute
budget. Three-dimensional lookup tables (3D LUTs) dominate real-time
enhancement because they decide at low resolution and apply a per-pixel lookup
at full resolution. A single global LUT, however, is spatially invariant, so an
underexposed shadow and a well-exposed region that share a pixel value receive
identical corrections. Spatially heterogeneous demands cannot be expressed by
such a mapping. We propose LoopLUT, a region-cascaded 3D LUT with progressive
refinement. A global LUT performs the overall correction, followed by $K-1$ loop
iterations. In each iteration a gating head predicts at low resolution the
region that still needs correction, then builds a residual LUT from the color
statistics of that region alone. The cascaded gates form a partition of unity,
so the output is a per-pixel convex combination of the $K$ lookup results.
Fusion is therefore performed by the gates themselves, with no separate fusion
module and no interpolation error accumulating across rounds. The decision stage
runs at a fixed $256\times256$ resolution, independent of output resolution, so
a 4K image costs only $K$ pure lookups. Extensive experiments across four
benchmarks show that LoopLUT improves PSNR by up to 2.81\,dB over the strongest
prior method, while keeping real-time throughput at 4K. The same decomposition
also generalizes well to underwater enhancement datasets.
\end{abstract}

\suppressfloats[t]

\section{Introduction}

\begin{figure}[t]
  \centering
  \includegraphics[width=0.88\linewidth]{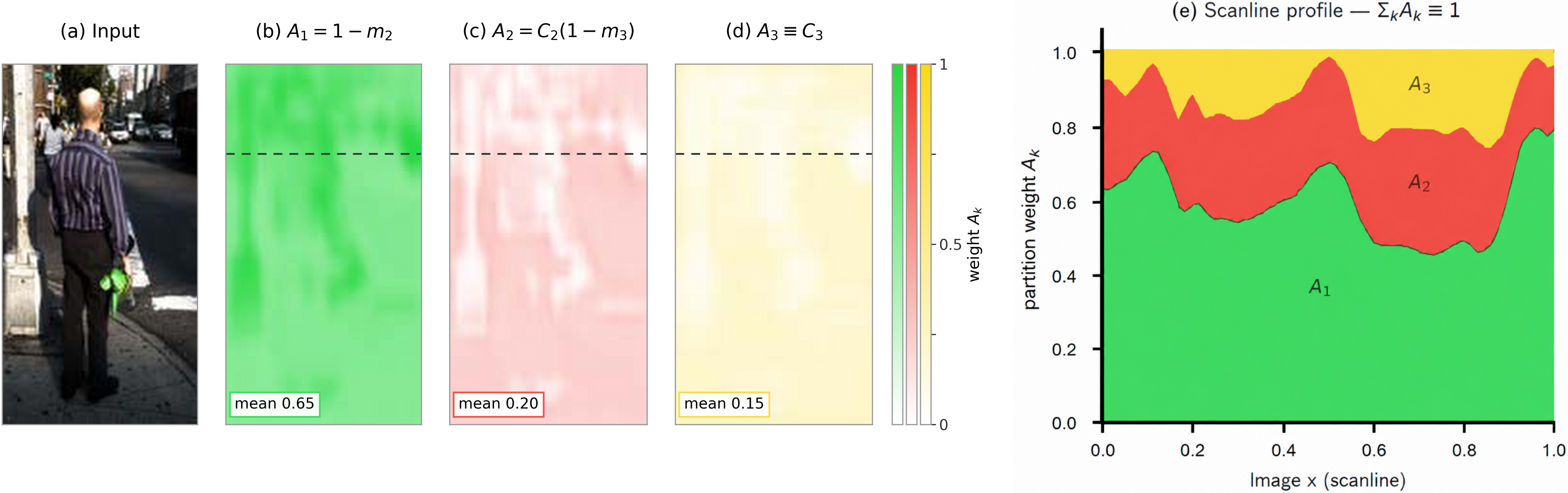}
  \caption{LoopLUT divides an image among its correction rounds. (a)~An LCDP
  test image. (b--d)~Partition weights from the trained gates of LoopLUT
  ($K{=}3$), shown without contrast stretching; color saturation encodes the
  weight. $\mathbf{A}_k(x)$ is the share of pixel $x$ taken from the lookup of
  round $k$. The means (0.65, 0.20 and 0.15) show that the global round covers
  most of the image and later rounds refine ever smaller regions.
  Each weight is what the cascade passes to round $k$ minus what round $k$ passes
  on, $\mathbf{A}_k{=}\mathbf{C}_k(1{-}\mathbf{m}_{k+1})$, where $\mathbf{C}_k$ is the product of the gates up to
  round $k$. The weights are therefore non-negative and sum to one at every
  pixel, so the output is a convex blend of the $K$ lookups. (e)~Profile along
  the dashed scanline, confirming the unit sum and smooth transitions.}
  \label{fig:teaser}
\end{figure}

Color enhancement of 4K images must preserve full-resolution detail within the
compute budget of mobile and video applications. The dominant real-time route
therefore decides at low resolution and executes at full resolution, as in
HDRNet~\citep{gharbi2017bilateral} and image-adaptive 3D
LUTs~\citep{zeng2020learning}. A 3D LUT suits this route, since applying it
costs one lookup per pixel.

This efficiency rests on a strong assumption, that enhancement is a single
global RGB-to-RGB mapping independent of spatial position and image content. Two
pixels with the same value therefore receive the same correction. Real
photographic degradation, however, varies within a scene. An
underexposed shadow needs brightening while an overexposed sky needs
suppression, even when the two contain similar pixel values. On mixed-exposure
scenes such as those of LCDP~\citep{wang2022lcdp}, a global LUT, however
adaptive, can only compromise between shadows and highlights.
Figure~\ref{fig:teaser} shows how our cascaded gates instead split such a scene
into differently corrected regions.

Existing work attacks this limitation from two directions, each leaving a gap.
(i)~Spatially aware operators add a weight map or local filters on top of
the lookup~\citep{wang2021spatial,gharbi2017bilateral,moran2020deeplpf}. Their
spatial decision is one-shot, predicted directly from the input, with no
feedback in which an applied enhancement determines the next correction. Several
also run the weight network at high resolution, so decision cost grows with
output resolution. (ii)~Iterative enhancement
instead approaches the target in multiple steps, as in diffusion-based
enhancement~\citep{jiang2024lightendiffusion}, flow
matching~\citep{hu2025flowlut} and recurrent optical-flow
refinement~\citep{teed2020raft}. In color enhancement it appears as repeated
curve application~\citep{guo2020zerodce}, LUT evolution~\citep{hu2025flowlut}
and recurrent state propagation~\citep{bai2026loopmamba}. Its value lies in
feedback, since each round is conditioned on the previous result and harder
regions can be deferred to later rounds. These methods, however, re-run the
entire network every round, so cost grows linearly with round count.
A second problem arises when several LUTs are applied in sequence, each to the
output of the previous one, and their results are fused by scalar weights. The
rounds are then nested, since the last already contains the effect of all
earlier ones, and fusion degenerates into picking the most complete round. Spatial
adaptivity, progressive per-round refinement, and a resolution-independent
decision cost have not yet held together in one architecture.

We propose LoopLUT to bridge this gap. Iteration helps through feedback rather
than through repeated full-resolution computation, so the loop should iterate
over inexpensive low-resolution decisions, namely where to correct and which
LUT to apply, rather than over full-resolution images. First, a global LUT
performs the overall correction. Second, in each later round, a lightweight
gating head predicts at low resolution, from the current enhancement state, the
region that still requires correction. Then, a residual LUT is generated from
the color statistics inside that region alone. Finally, the per-round gates
multiply into a cascade that forms a partition of unity over the image, so the
output is a per-pixel convex combination of the $K$ lookup results. Because the
gates themselves perform the fusion, no separate module exists for a trivial
solution to bypass. All $K$ LUTs also act directly on the original image,
eliminating the compounded interpolation error of serial composition.

Our contributions are as follows.
\begin{itemize}
  \item Region-local color correction through cascaded gates. The gates form a
  partition of unity and perform the fusion themselves, so the effective LUT
  varies per pixel and a shadow and a well-exposed region sharing a pixel value
  receive different corrections.
  \item A low-resolution decision loop. Each gate is predicted from the result
  of the previous round, so later rounds refine only what earlier rounds left,
  while the loop stays at a constant $256\times256$ and full-resolution work
  remains a pure lookup.
  \item A deployed system, not only a benchmark result. We export
  LoopLUT to ONNX and run it on an Android phone at about 100\,--\,200\,ms per
  frame on CPU alone (Appendix~\ref{sec:mobile}), and validate cross-task
  generality on four public benchmarks.
\end{itemize}

\section{Related Work}

\paragraph{3D LUT enhancement.}
\citet{zeng2020learning} brought learnable 3D LUTs into the
low-resolution-decision paradigm of HDRNet~\citep{gharbi2017bilateral} and
CSRNet~\citep{he2020conditional}, predicting basis-LUT coefficients so that one
full-resolution lookup suffices. Later work refines the table through non-uniform lattices~\citep{yang2022adaint,zhou2026selfdistilled},
factorized or compressed tables~\citep{yang2022seplut,zhang2022clutnet},
implicit neural fields~\citep{conde2024nilut}, attention-based
fusion~\citep{fu2024attentionlut} and folded networks~\citep{yang2024taming},
but the operator remains global.

\paragraph{Local enhancement.}
Local methods apply region-specific corrections through spatially aware weight
maps~\citep{wang2021spatial}, context-aware 4D LUTs~\citep{liu20234dlut},
per-pixel lookup weights~\citep{liu2024pixellut}, local parametric
filters~\citep{moran2020deeplpf}, multi-scale pyramids~\citep{afifi2021exposure}
or local color distribution priors~\citep{wang2022lcdp}. Their regional decision
is one-shot and predicted from the input alone, so a missed or over-corrected
region is never revisited. Several also run the weight network at high
resolution.

\paragraph{Iterative refinement.}
Many models reach their target through conditioned rounds, from recurrent
optical flow~\citep{teed2020raft}, repeated curve
application~\citep{guo2020zerodce} and unrolled Retinex
optimization~\citep{liu2021ruas} to multi-step
diffusion~\citep{jiang2024lightendiffusion}, flow-matching LUT
evolution~\citep{hu2025flowlut} and recurrent restoration~\citep{bai2026loopmamba}.
Their strength is feedback. Each round sees the previous result, so it corrects
the remaining error rather than predicting the whole mapping at once. These
loops, however, operate on the image or its feature maps, so every round repeats
image-sized computation. LoopLUT moves the loop into the decision stage. Each
round conditions its gate and residual LUT on a low-resolution rehearsal of the
previous result (Section~\ref{sec:gate}), so later rounds address only what
earlier rounds left, at the cost of a 16\,K-parameter gating head and one MLP at
$256\times256$. Predicting regions round by round also makes locality part of
the loop rather than a module added to a global operator.

\FloatBarrier

\section{Method}

\subsection{Notation and overall framework}

\begin{figure}[t]
  \centering
  \includegraphics[width=0.92\linewidth]{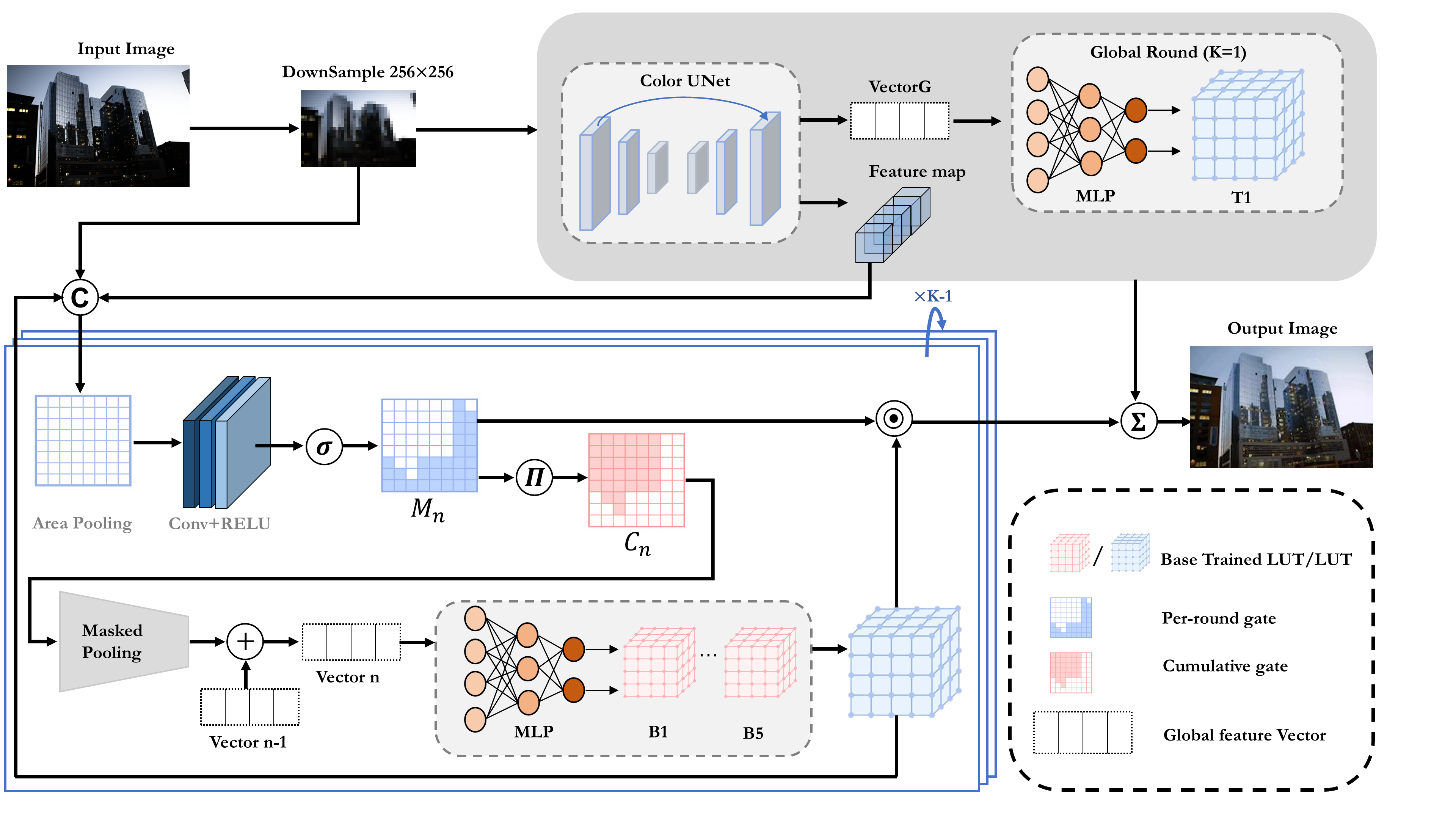}
  \caption{Overall architecture of LoopLUT ($\mathbf{M}_n$, $\mathbf{C}_n$ and $\mathbf{T}_n$ in the figure
  are $\mathbf{m}_k$, $\mathbf{C}_k$ and $\mathbf{T}_k$ in the text). The global round (gray)
  passes the $256\times 256$ input once through the Color UNet to give the
  global feature $\mathbf{F}_{\text{g}}$ (VectorG) and the decoder feature map $\mathbf{D}$ (Feature map),
  which an MLP turns into the global LUT $\mathbf{T}_1$ by mixing the basis LUTs
  $\mathbf{B}_1,\dots,\mathbf{B}_5$. Each region round (blue, $\times K{-}1$) area-pools
  its inputs to the $s\times s$ gate bottleneck, predicts the gate $\mathbf{M}_n$,
  multiplies it into the cumulative gate $\mathbf{C}_n$, and generates $\mathbf{T}_n$ from masked
  statistics inside $\mathbf{C}_n$. At execution the $K$ LUTs are applied to the
  original image and combined per pixel by the partition weights, so the gates
  are the fusion. The decision stage is fixed at low resolution throughout.}
  \label{fig:arch}
\end{figure}

Given a degraded input $\mathbf{I}\in[0,1]^{3\times H\times W}$, the goal is an enhanced
image $\hat{\mathbf{I}}$ of the same size. A 3D LUT is a grid
$\mathbf{T}\in\mathbb{R}^{3\times S\times S\times S}$ on the RGB cube (here $S{=}33$)
acting by per-pixel trilinear interpolation,
$\mathbf{T}(\mathbf{I})(x)=\trilerp\bigl(\mathbf{T};\,\mathbf{I}(x)\bigr)$, which is independent of the pixel
coordinate $x$. LoopLUT composes $K$ such LUTs (here $K{=}3$) with $K{-}1$
spatial gates into a spatially varying mapping, while keeping full-resolution
computation to lookups alone.

The decision stage (Figure~\ref{fig:arch}) downsamples the input to
$\Ilo$ at a fixed $256\times 256$ and passes it through a UNet
backbone~\citep{ronneberger2015unet} with GroupNorm~\citep{wu2018groupnorm},
yielding a global color feature $\mathbf{F}_{\text{g}}$ and a decoder feature map $\mathbf{D}$. The first
round generates the global LUT $\mathbf{T}_1$. Each round $k=2,\dots,K$ then predicts a
gate $\mathbf{m}_k$, cascades it into the cumulative gate $\mathbf{C}_k$, pools color statistics
inside $\mathbf{C}_k$ to generate a region residual $\boldsymbol{\Delta}_k$, and rehearses the current
enhancement state at low resolution. One backbone pass, $K{-}1$ gating heads and
$K$ MLP evaluations all run at $\le 256\times 256$, so the decision cost is
independent of the output resolution. The execution stage performs $K$
pure lookups on the original image and combines them per pixel with the
bilinearly upsampled partition weights, so its cost grows only linearly in pixel
count.

\subsection{Region cascade: the gates are the fusion}
\label{sec:cascade}

Each round's gate $\mathbf{m}_k\in(0,1)$ indicates whether a pixel still requires the
$k$-th round of correction. With the cumulative gate $\mathbf{C}_k=\prod_{j=2}^{k}\mathbf{m}_j$,
the partition weights and the output are
\begin{equation}
  \mathbf{A}_1 = 1-\mathbf{m}_2,\qquad
  \mathbf{A}_k = \mathbf{C}_k\,(1-\mathbf{m}_{k+1}),\ \ 1<k<K,\qquad
  \mathbf{A}_K = \mathbf{C}_K,
  \label{eq:partition}
\end{equation}
\begin{equation}
  \hat{\mathbf{I}}(x) = \sum_{k=1}^{K} \mathbf{A}_k(x)\, \mathbf{T}_k(\mathbf{I})(x).
  \label{eq:output}
\end{equation}
The telescoping structure of Equation~\eqref{eq:partition} gives three
properties, proved in Appendix~\ref{sec:appendix_proof}. First,
$\sum_k \mathbf{A}_k\equiv 1$ pointwise with every $\mathbf{A}_k$ non-negative, so the output is a
per-pixel convex combination of the $K$ lookup results and cannot leave the
valid range. Second, the supports contract monotonically,
$\supp(\mathbf{A}_K)\subseteq\supp(\mathbf{C}_K)\subseteq\cdots\subseteq\supp(\mathbf{C}_2)$, so each later
round refines only inside the region the previous round left marked as needing
correction. Third, substituting the residual form
$\mathbf{T}_k=\mathbf{T}_1+\sum_{j=2}^{k}\boldsymbol{\Delta}_j$ and using the linearity of trilinear
interpolation gives the exact identity
\begin{equation}
  \hat{\mathbf{I}}(x) = \mathbf{T}^{x}_{\text{eff}}(\mathbf{I})(x),\qquad
  \mathbf{T}^{x}_{\text{eff}} = \mathbf{T}_1 + \sum_{k=2}^{K} \mathbf{C}_k(x)\,\boldsymbol{\Delta}_k,
  \label{eq:efflut}
\end{equation}
so the model is exactly a per-pixel effective LUT: a global component plus
region residuals whose coefficients are the cumulative gates. Equation~\eqref{eq:output}
has no separate fusion step, since the partition weights are the gates
themselves, and all $K$ LUTs act on the original image, removing the
compounded interpolation error of serial composition.
Appendix~\ref{sec:appendix_design} quantifies both defects.

\subsection{Recurrent decision with progressive region refinement}
\label{sec:gate}

Where the $k$-th round should refine depends on the gap remaining between the
current result and the ideal one, which is unavailable at inference. LoopLUT
estimates no error quantity; the gating head instead learns end to end, from
inference-time inputs, which regions still need correcting:
\begin{equation}
  \mathbf{m}_k = \Bigl(\sigma\bigl(g_k\bigl(\downarrow_s
  [\,\mathbf{D},\ \Ilo,\ \mathbf{q}_{k-1},\ |\mathbf{q}_{k-1}-\Ilo|\,]\bigr)\bigr)\Bigr)\uparrow,
  \label{eq:gate}
\end{equation}
where $g_k$ is a three-layer convolutional network, independent per round (about
16\,K parameters), $\downarrow_s$ is downsampling by area averaging to
$s\times s$ (here $s{=}32$), and $\uparrow$ is bilinear upsampling back to the
decision resolution. The bottleneck $s$ is a structural constraint rather than
an approximation for saving computation, and it bounds the spatial precision of
the gate (Appendix~\ref{sec:appendix_design}).

The residual is then driven by the statistics of the region it actually faces.
Masked weighted pooling inside the cumulative gate gives $\mathbf{F}_k = \mathbf{F}_{\text{g}} + P(\mathbf{z}_k)$,
where $\mathbf{z}_k$ collects the gate-weighted means $\mu_{\mathbf{C}_k}(\mathbf{D})$, $\mu_{\mathbf{C}_k}(\Ilo)$
and $\mu_{\mathbf{C}_k}(|\mathbf{q}_{k-1}-\Ilo|)$ together with the area fraction $\bar{\mathbf{C}}_k$, with
${\mu_{\mathbf{C}}(\cdot)=\sum(\mathbf{C}\odot\cdot)/(\sum \mathbf{C}+\varepsilon)}$ and $P$ a two-layer MLP
whose last layer is zero-initialized.

Closing the feedback at full resolution would destroy the efficiency skeleton,
so the current enhancement state is kept inside the decision stage by a
low-resolution rehearsal, $\mathbf{q}_1=\mathbf{T}_1(\Ilo)$ and
$\mathbf{q}_k=\mathbf{q}_{k-1}+\mathbf{C}_k\odot\bigl(\mathbf{T}_k(\Ilo)-\mathbf{T}_{k-1}(\Ilo)\bigr)$ for $k\ge 2$, which
replays the execution stage on $\Ilo$ under the same cascade of weights. Since
$\mathbf{q}_k$ feeds the next round's gate and region statistics, every decision is
conditioned on the state after the previous round's correction while all
feedback computation stays at $256\times 256$.

\subsection{LUT generation and training objective}
\label{sec:lutgen}
\label{sec:loss}

Each round updates $\mathbf{T}_k=\text{clip}(\mathbf{T}_{k-1}+\gamma\sum_{i=1}^{n}c_{k,i}\mathbf{B}_i)$ from the identity LUT
$\mathbf{T}_0$ by mixing basis LUTs~\citep{zeng2020learning}, with residual
$\boldsymbol{\Delta}_k=\mathbf{T}_k-\mathbf{T}_{k-1}$ and
$\mathbf{c}_k=\text{MLP}\bigl(\mathbf{F}_k\oplus E(k)\bigr)$, where $\{\mathbf{B}_i\}$ are $n{=}5$
learnable basis LUTs (about 0.54\,M parameters), $E(k)$ is a sinusoidal round
encoding and $\gamma$ is an overall scale. Basis mixing uses two orders of
magnitude fewer parameters than regressing the $3S^3$-dimensional grid directly
and carries an inherent low-rank constraint.

The objective is
$\mathcal{L}=\mathcal{L}_{\text{rec}}+\lambda_{\text{a}}\mathcal{L}_{\text{area}}
+\lambda_{\text{tv}}\mathcal{L}_{\text{tv}}+\lambda_{\text{s}}\mathcal{L}_{\text{smooth}}
+\lambda_{\text{g}}\mathcal{L}_{\text{gate}}$. The reconstruction term
$\mathcal{L}_{\text{rec}}$ combines $\ell_1$, SSIM~\citep{wang2004ssim} and
contrast and saturation regularizers on the final output $\hat{\mathbf{I}}$; by
Equation~\eqref{eq:output} every round of LUTs and gates is supervised through
that single path, so no per-round intermediate supervision is needed. The region
coverage term $\mathcal{L}_{\text{area}}$ is a two-sided hinge on the mean
cumulative gate $\bar{\mathbf{C}}_k$, with a per-round decreasing upper bound $\tau_k$
that carries the inductive bias of contraction and a small floor $\tau_{\min}$
that keeps the gradient path open. $\mathcal{L}_{\text{tv}}$ and
$\mathcal{L}_{\text{smooth}}$ smooth the gates and the LUTs, the latter with
an added monotonicity hinge against color banding, and
$\mathcal{L}_{\text{gate}}$ supplies an auxiliary gate target that exists only
during training. Appendix~\ref{sec:appendix_design} gives these terms in full,
together with the initialization constraint imposed by the product structure and
the reason a coverage bound is used in place of an error threshold.

\section{Experiments}
\label{sec:exp}

\subsection{Datasets and metrics}
\label{sec:setup}

We evaluate on four public benchmarks: LCDP~\citep{wang2022lcdp} for mixed
exposure, MIT-Adobe FiveK~\citep{bychkovsky2011fivek} for expert retouching,
Mobile-Spec~\citep{zhou2024mobilespec} for mobile photography, and
LSUI~\citep{peng2023ushape} for underwater enhancement. LCDP is our primary
benchmark and carries all ablations. We report PSNR, SSIM~\citep{wang2004ssim},
LPIPS~\citep{zhang2018lpips} with the AlexNet backbone, and CIEDE2000, all
evaluated at native resolution without cropping or rescaling. Splits and the
full evaluation protocol are given in Appendix~\ref{sec:appendix_data}.

\subsection{Implementation details}
\label{sec:impl}

All experiments are run on a single NVIDIA RTX 4050 Laptop GPU (6\,GB). During
training the whole input image is resized to $256\times 256$ rather than
randomly cropped, because LUT generation depends on the color statistics of the
entire image and cropping would make the framing protocol inconsistent between
training and inference. The structural hyperparameters match those in the method
section: $K{=}3$ rounds, gate bottleneck $s{=}32$, $n{=}5$ basis LUTs and LUT
side length $S{=}33$. The model is trained for 100 epochs.

\paragraph{Compared methods.} We compare against nine methods:
HDRNet~\citep{gharbi2017bilateral}, RUAS~\citep{liu2021ruas},
LCDPNet~\citep{wang2022lcdp}, AdaInt~\citep{yang2022adaint},
LightenDiffusion~\citep{jiang2024lightendiffusion}, CSEC~\citep{li2024csec},
DnLUT~\citep{yang2025dnlut}, CanonCGT~\citep{ko2026canoncgt} and
ShiftLUT~\citep{zeng2026shiftlut}; the reasons for this selection are given in
Appendix~\ref{sec:appendix_baselines}. On LSUI we compare against methods
designed for underwater enhancement (Table~\ref{tab:lsui}).

\begin{table}[t]
  \centering
  \caption{Quantitative comparison with state-of-the-art methods on LCDP,
  MIT-Adobe FiveK and Mobile-Spec. In each column the \textbf{best} is bold and
  the \underline{second best} is underlined; $\uparrow$/$\downarrow$ indicates
  that higher/lower is better.}
  \label{tab:main}
  \setlength{\tabcolsep}{3pt}
  \resizebox{\linewidth}{!}{%
  \begin{tabular}{@{}lccccccccccccc@{}}
    \toprule
    & & \multicolumn{4}{c}{LCDP} & \multicolumn{4}{c}{MIT-Adobe FiveK}
      & \multicolumn{4}{c}{Mobile-Spec} \\
    \cmidrule(lr){3-6}\cmidrule(lr){7-10}\cmidrule(lr){11-14}
    Method & Venue
    & PSNR$\uparrow$ & SSIM$\uparrow$ & LPIPS$\downarrow$ & CIEDE$\downarrow$
    & PSNR$\uparrow$ & SSIM$\uparrow$ & LPIPS$\downarrow$ & CIEDE$\downarrow$
    & PSNR$\uparrow$ & SSIM$\uparrow$ & LPIPS$\downarrow$ & CIEDE$\downarrow$ \\
    \midrule
    HDRNet & SIGGRAPH'17
    & 21.75 & 0.8265 & 0.1860 & 10.98
    & 20.57 & \underline{0.7251} & \underline{0.0960} & 9.10
    & 20.34 & 0.7911 & 0.1652 & 6.57 \\
    RUAS & CVPR'21
    & 18.25 & 0.7568 & 0.2065 & 9.11
    & 19.29 & 0.7095 & 0.1393 & 9.86
    & 12.22 & 0.7470 & 0.1833 & 17.26 \\
    LCDPNet & ECCV'22
    & 23.29 & 0.8423 & 0.1587 & 8.43
    & 15.13 & 0.6029 & 0.2437 & 17.17
    & 25.25 & 0.9097 & 0.1404 & 5.21 \\
    AdaInt & CVPR'22
    & 20.93 & 0.7856 & 0.2258 & 9.42
    & 18.76 & 0.7235 & 0.1474 & 12.21
    & \underline{26.98} & \underline{0.9264} & 0.0850 & \underline{4.02} \\
    LightenDiffusion & ECCV'24
    & 19.04 & 0.7743 & \underline{0.1334} & 9.33
    & 18.35 & 0.7050 & 0.1542 & 12.35
    & 23.23 & 0.8383 & 0.2759 & 7.31 \\
    CSEC & CVPR'24
    & \underline{23.35} & \underline{0.8552} & 0.1563 & 7.93
    & 12.88 & 0.5668 & 0.2757 & 21.13
    & 24.13 & 0.9023 & 0.1565 & 5.79 \\
    DnLUT & CVPR'25
    & 21.02 & 0.7563 & 0.3002 & 8.68
    & \underline{20.77} & 0.7119 & 0.2431 & \underline{9.05}
    & 25.19 & 0.8716 & 0.2068 & 5.30 \\
    CanonCGT & CVPR'26
    & 22.92 & 0.8241 & \textbf{0.1264} & \textbf{7.03}
    & 19.37 & 0.7030 & 0.2086 & 14.15
    & 25.04 & 0.9254 & \underline{0.0701} & 4.83 \\
    ShiftLUT & CVPR'26
    & 20.25 & 0.7419 & 0.2412 & 10.45
    & 19.67 & 0.7122 & 0.1357 & 10.50
    & 24.32 & 0.8824 & 0.1893 & 7.02 \\
    \midrule
    LoopLUT (ours) & --
    & \textbf{23.48} & \textbf{0.8616} & 0.1558 & \underline{7.53}
    & \textbf{21.92} & \textbf{0.7501} & \textbf{0.0875} & \textbf{7.79}
    & \textbf{29.79} & \textbf{0.9680} & \textbf{0.0458} & \textbf{2.91} \\
    \bottomrule
  \end{tabular}}
\end{table}

\subsection{Comparison with state-of-the-art methods}
\label{sec:comparison}
\label{sec:qualitative}

Table~\ref{tab:main} shows that LoopLUT ranks first on all four metrics on
MIT-Adobe FiveK and Mobile-Spec. Detailed metrics averaged over the three benchmarks are
given in Appendix~\ref{sec:appendix_avg} (Table~\ref{tab:avg} and Figure~\ref{fig:avg}). Its PSNR exceeds the runner-up by 1.15\,dB on
FiveK (DnLUT, 20.77\,dB) and by 2.81\,dB on Mobile-Spec (AdaInt, 26.98\,dB). On
LCDP it ranks first on PSNR and SSIM (23.48\,dB and 0.8616), ahead of CSEC and
LCDPNet, which are designed for exposure correction. CanonCGT leads on the
remaining two metrics, CIEDE2000 (7.03 vs.\ 7.53) and LPIPS (0.1264 vs.\
0.1558), on which LoopLUT ranks second and third respectively.

LoopLUT is also the most consistent across datasets. Methods tailored to one
degradation drop sharply outside it: CSEC and LCDPNet reach only 12.88 and
15.13\,dB on FiveK, and RUAS 12.22\,dB on Mobile-Spec. Averaged over the three
datasets, LoopLUT ranks first on all four metrics, with 25.06\,dB PSNR against
22.44\,dB for the runner-up CanonCGT.

\begin{figure}[t]
  \centering
  \setlength{\tabcolsep}{1pt}
  \begin{tabular}{cccccc}
    \includegraphics[width=0.16\linewidth]{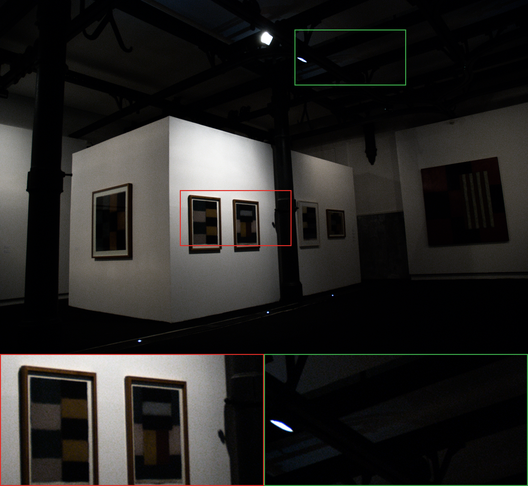} &
    \includegraphics[width=0.16\linewidth]{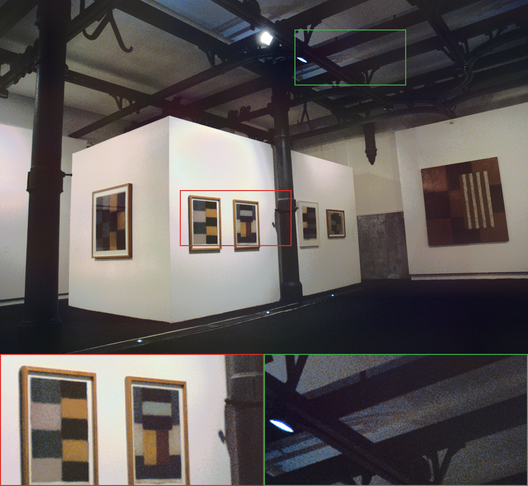} &
    \includegraphics[width=0.16\linewidth]{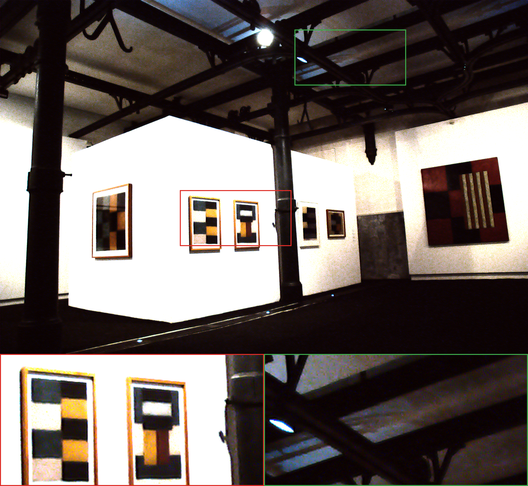} &
    \includegraphics[width=0.16\linewidth]{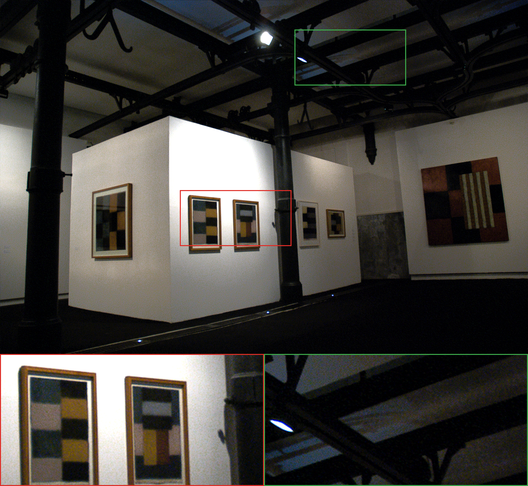} &
    \includegraphics[width=0.16\linewidth]{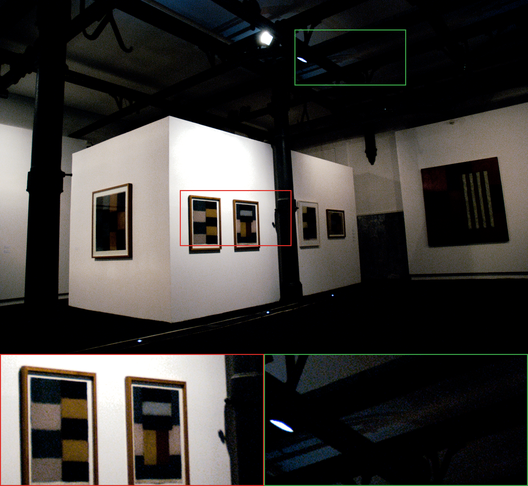} &
    \includegraphics[width=0.16\linewidth]{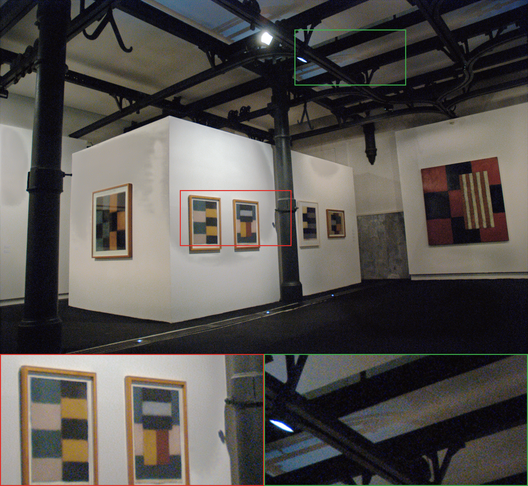} \\
    {\footnotesize Input} & {\footnotesize HDRNet} & {\footnotesize RUAS} &
    {\footnotesize LCDPNet} & {\footnotesize AdaInt} & {\footnotesize CSEC} \\
    \includegraphics[width=0.16\linewidth]{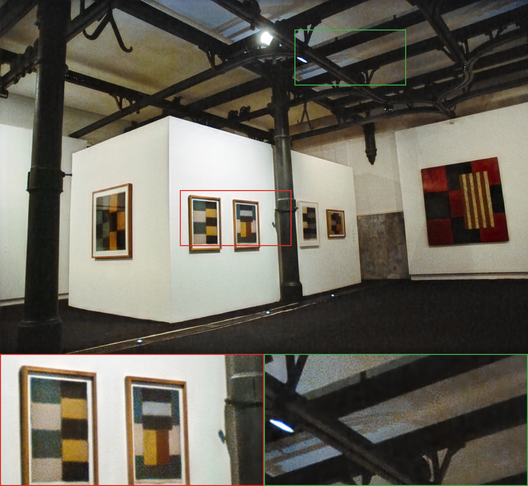} &
    \includegraphics[width=0.16\linewidth]{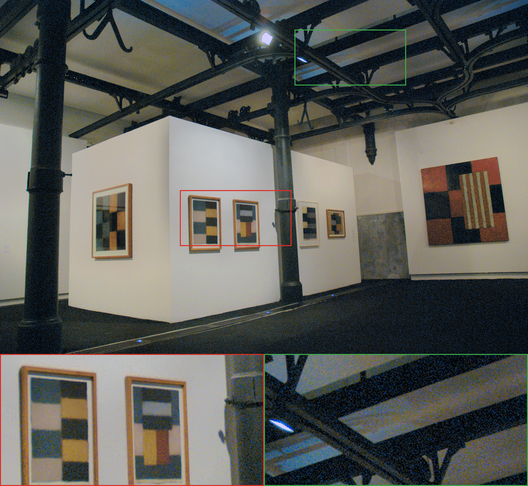} &
    \includegraphics[width=0.16\linewidth]{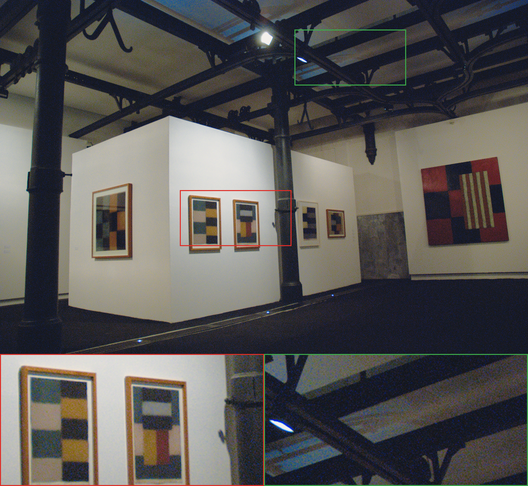} &
    \includegraphics[width=0.16\linewidth]{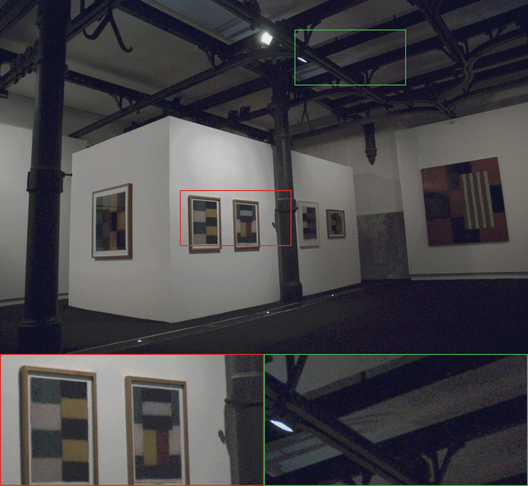} &
    \includegraphics[width=0.16\linewidth]{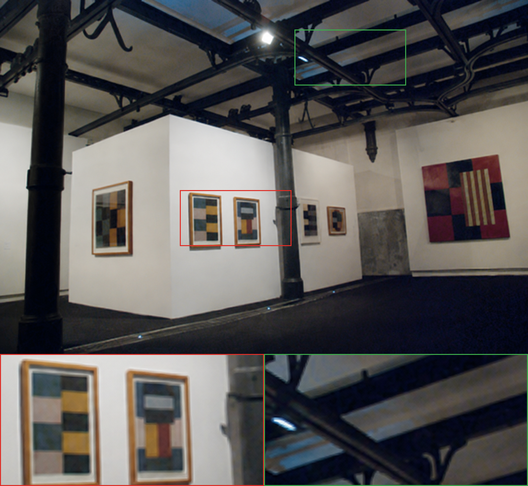} &
    \includegraphics[width=0.16\linewidth]{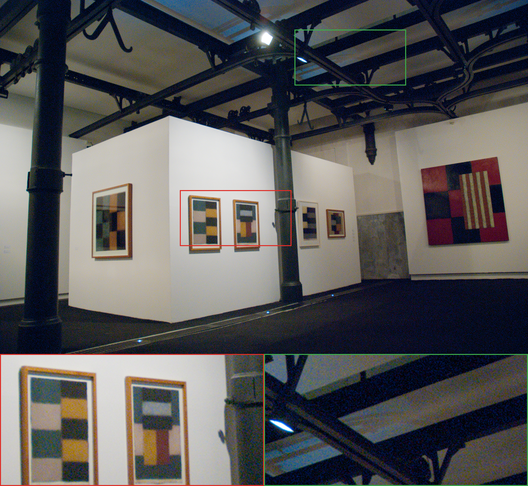} \\
    {\footnotesize LightenDiffusion} & {\footnotesize DnLUT} & {\footnotesize CanonCGT} &
    {\footnotesize ShiftLUT} & {\footnotesize LoopLUT (ours)} & {\footnotesize GT} \\
  \end{tabular}
  \caption{Qualitative comparison on the LCDP test set. Each panel shows the
  full output above and, below, the magnified red and green crops, placed on the
  two middle paintings and on the steel frame at the top; the red box examines
  midtone color reproduction and the green box detail recovery in dark
  structure. All methods use the same crops.}
  \label{fig:qual_lcdp}
\end{figure}

\begin{figure}[t]
  \centering
  \setlength{\tabcolsep}{1pt}
  \begin{tabular}{cccccc}
    \includegraphics[width=0.16\linewidth]{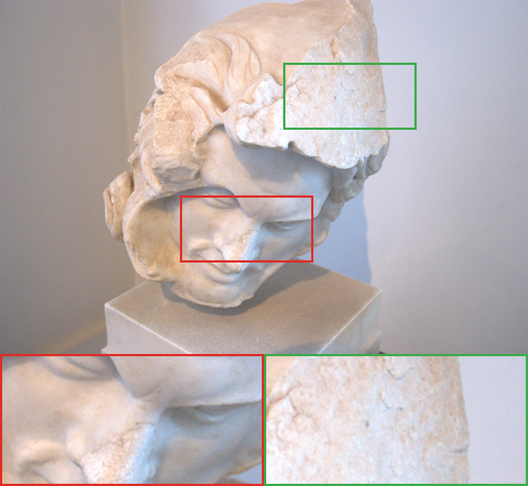} &
    \includegraphics[width=0.16\linewidth]{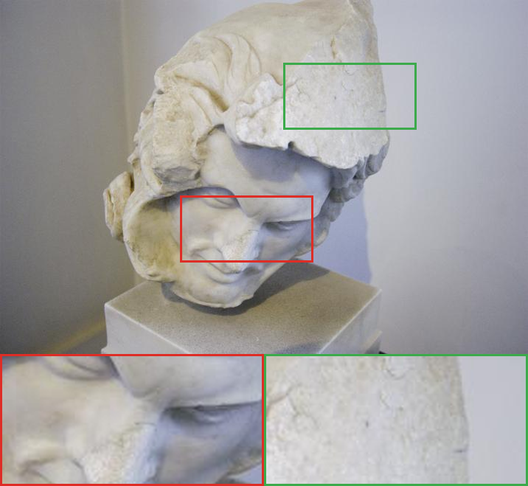} &
    \includegraphics[width=0.16\linewidth]{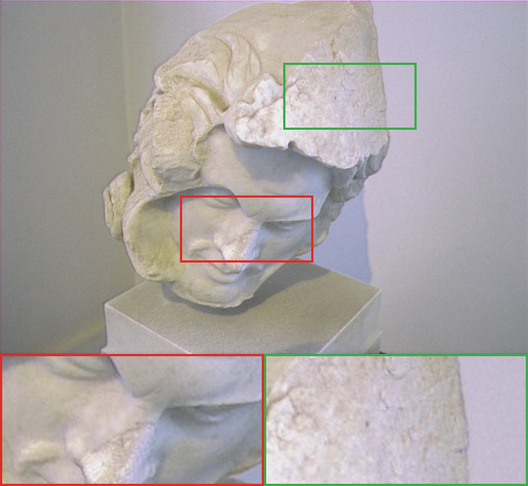} &
    \includegraphics[width=0.16\linewidth]{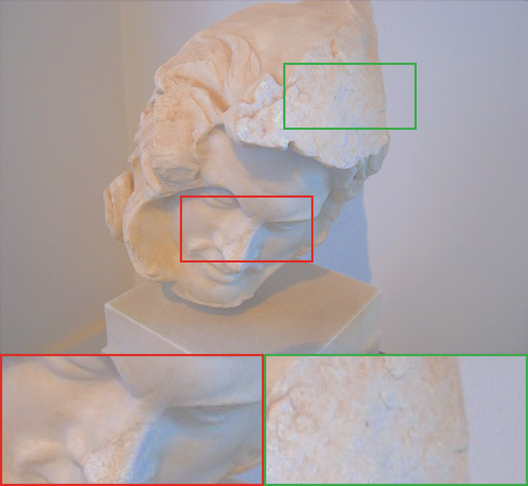} &
    \includegraphics[width=0.16\linewidth]{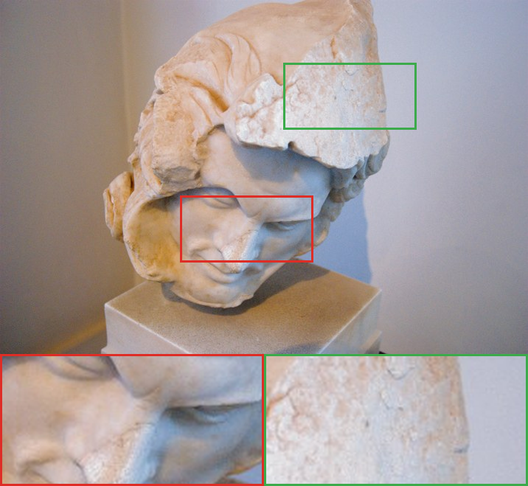} &
    \includegraphics[width=0.16\linewidth]{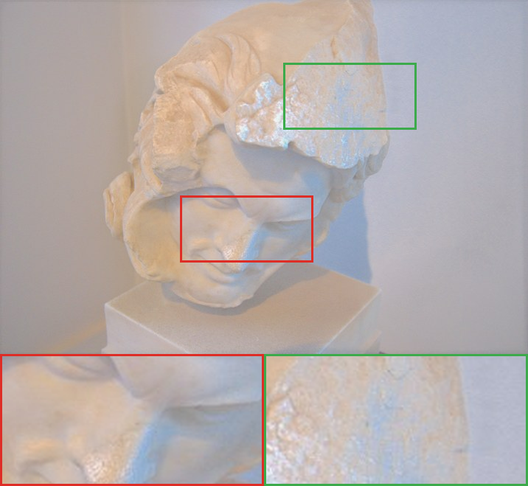} \\
    {\footnotesize Input} & {\footnotesize HDRNet} & {\footnotesize RUAS} &
    {\footnotesize LCDPNet} & {\footnotesize AdaInt} & {\footnotesize CSEC} \\
    \includegraphics[width=0.16\linewidth]{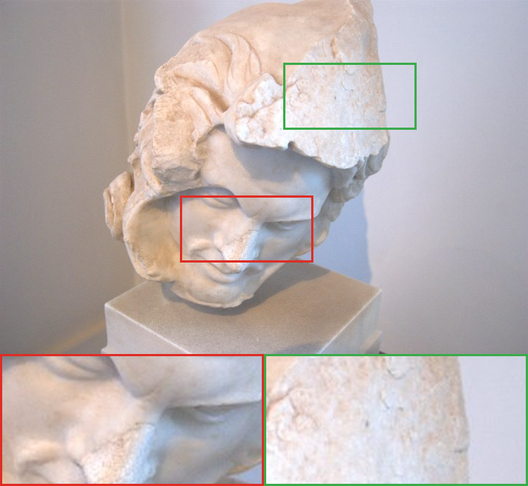} &
    \includegraphics[width=0.16\linewidth]{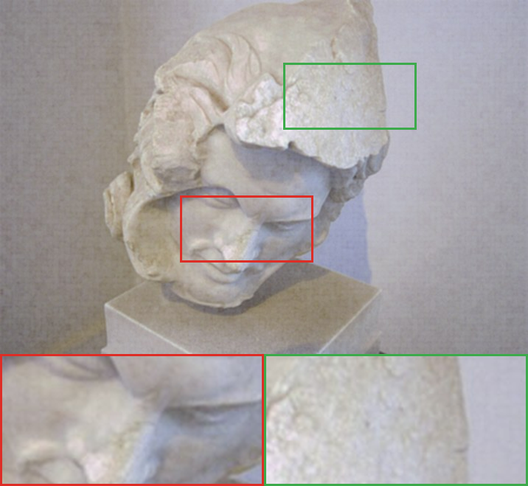} &
    \includegraphics[width=0.16\linewidth]{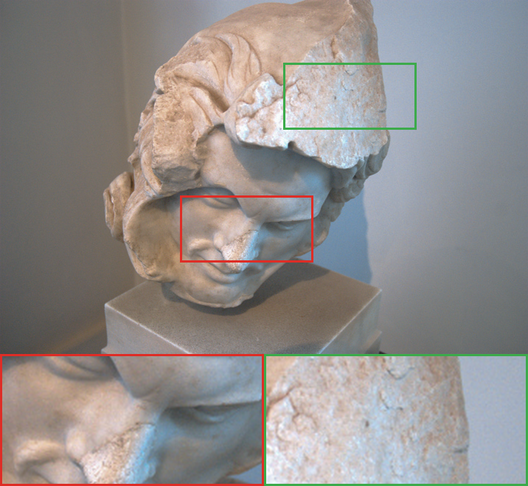} &
    \includegraphics[width=0.16\linewidth]{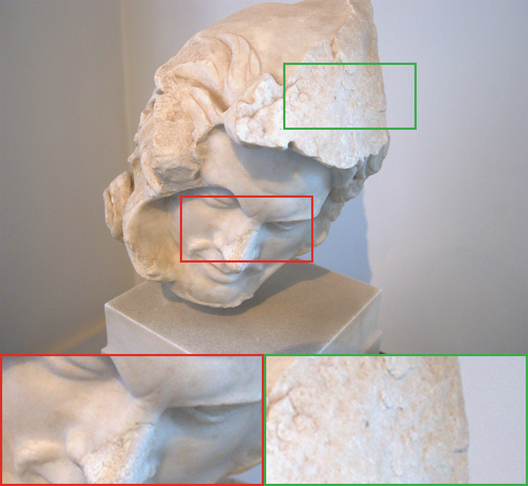} &
    \includegraphics[width=0.16\linewidth]{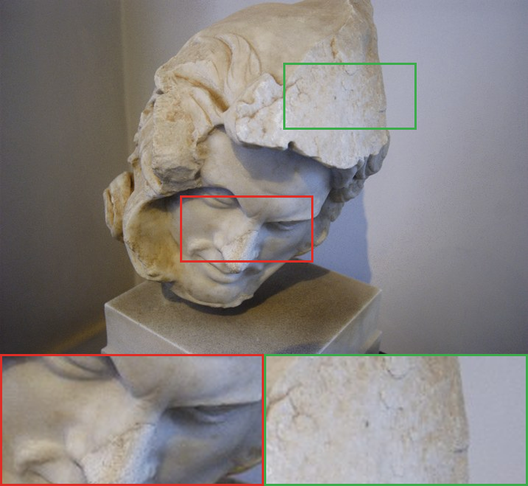} &
    \includegraphics[width=0.16\linewidth]{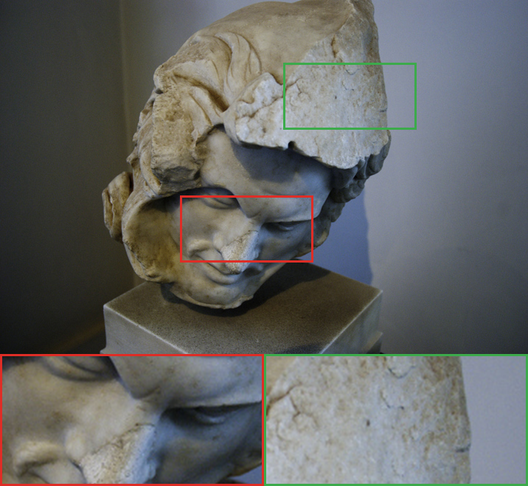} \\
    {\footnotesize LightenDiffusion} & {\footnotesize DnLUT} & {\footnotesize CanonCGT} &
    {\footnotesize ShiftLUT} & {\footnotesize LoopLUT (ours)} & {\footnotesize GT} \\
  \end{tabular}
  \caption{Qualitative comparison on the MIT-Adobe FiveK test set. Layout as in
  Figure~\ref{fig:qual_lcdp}; the red crop covers the face of a stone head and
  the green crop the weathered stone of its crown, examining the light-to-dark
  transition across the face and texture preservation on the highlight side.}
  \label{fig:qual_fivek}
\end{figure}

\begin{figure}[t]
  \centering
  \setlength{\tabcolsep}{1pt}
  \begin{tabular}{cccccc}
    \includegraphics[width=0.16\linewidth]{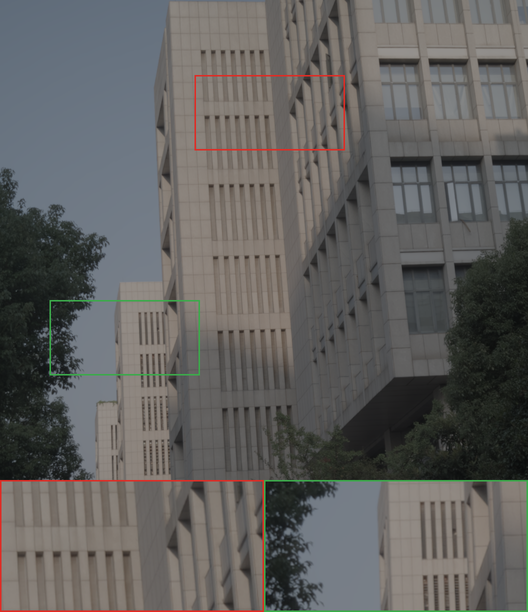} &
    \includegraphics[width=0.16\linewidth]{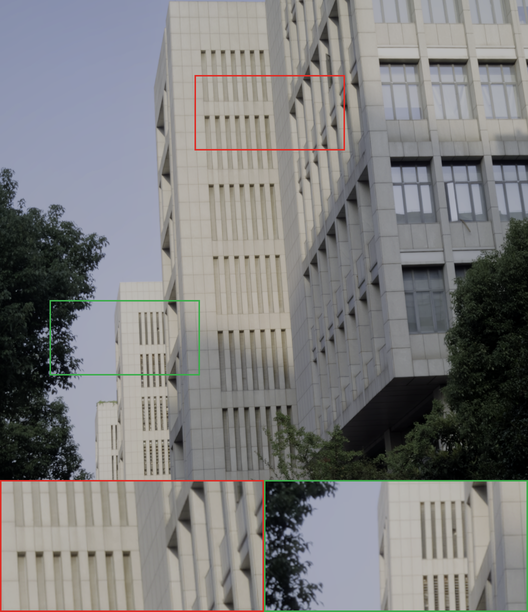} &
    \includegraphics[width=0.16\linewidth]{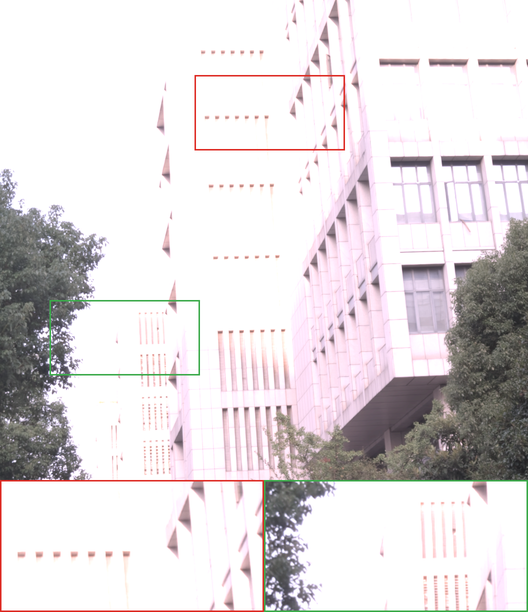} &
    \includegraphics[width=0.16\linewidth]{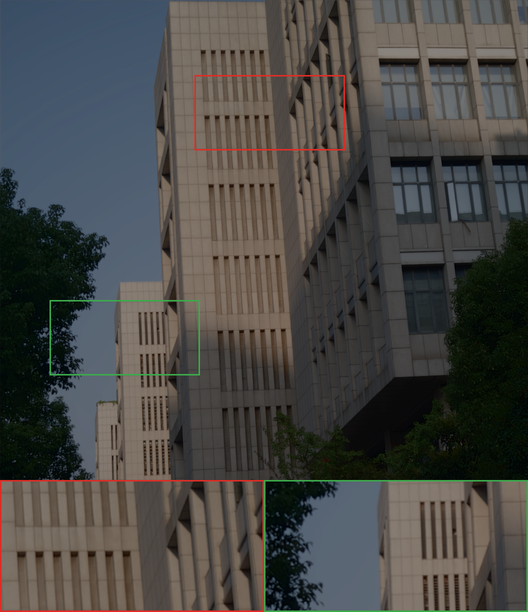} &
    \includegraphics[width=0.16\linewidth]{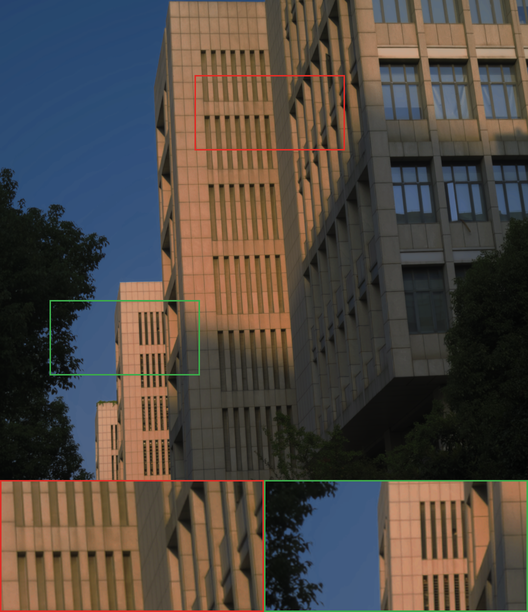} &
    \includegraphics[width=0.16\linewidth]{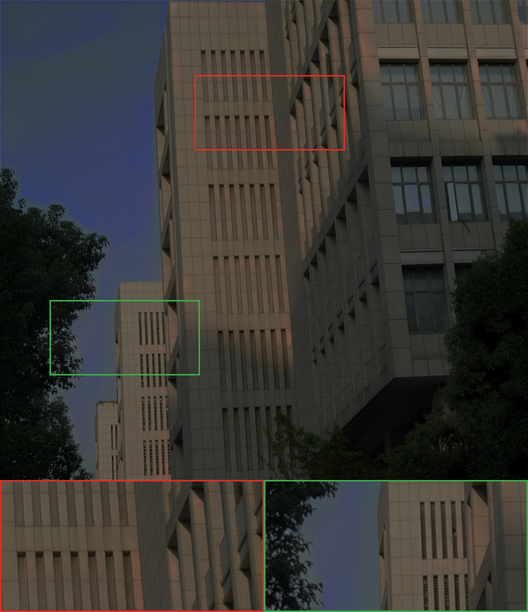} \\
    {\footnotesize Input} & {\footnotesize HDRNet} & {\footnotesize RUAS} &
    {\footnotesize LCDPNet} & {\footnotesize AdaInt} & {\footnotesize CSEC} \\
    \includegraphics[width=0.16\linewidth]{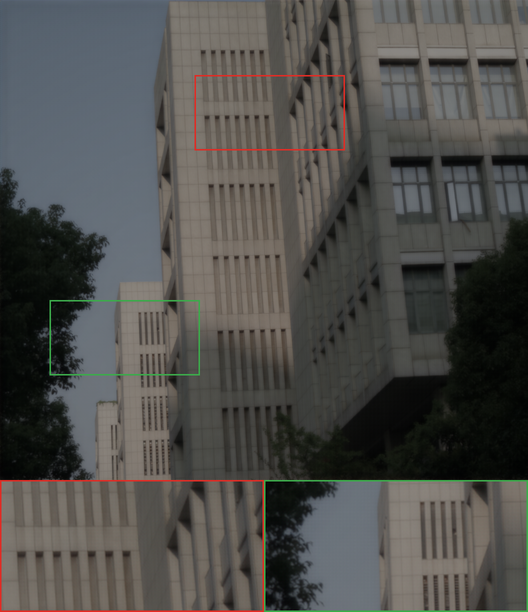} &
    \includegraphics[width=0.16\linewidth]{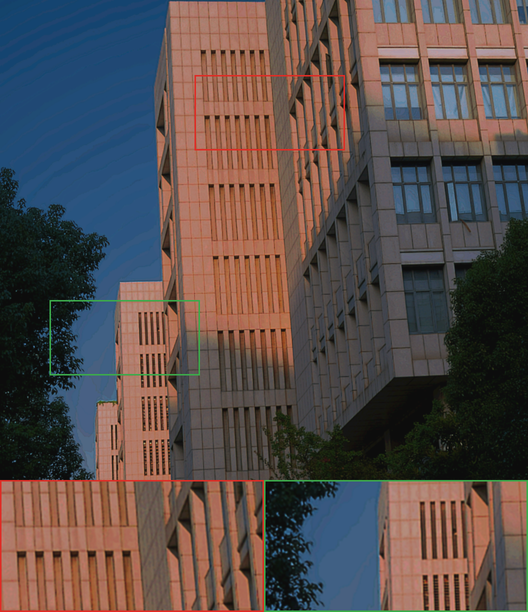} &
    \includegraphics[width=0.16\linewidth]{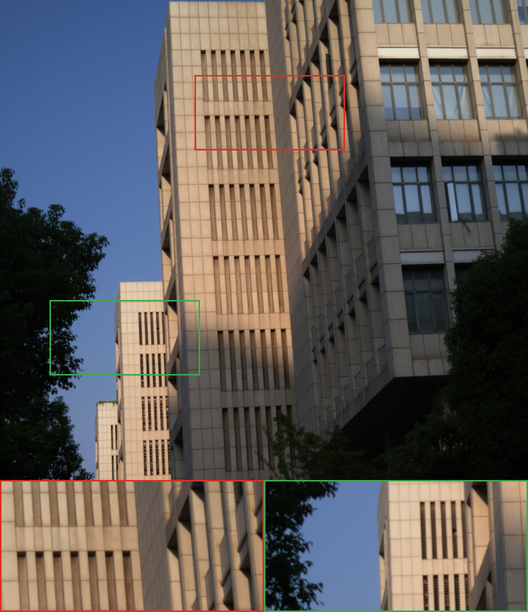} &
    \includegraphics[width=0.16\linewidth]{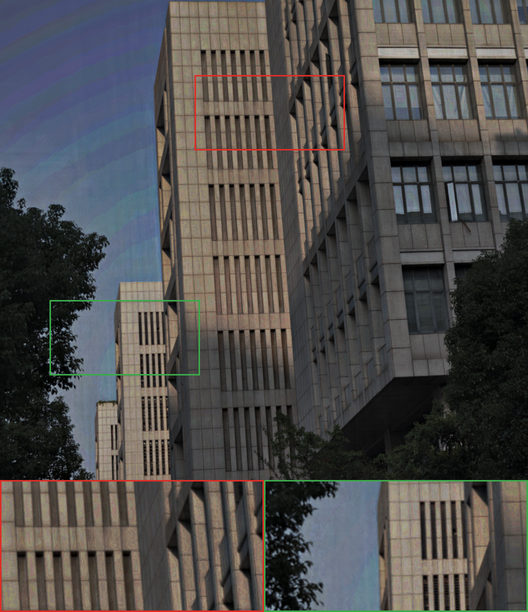} &
    \includegraphics[width=0.16\linewidth]{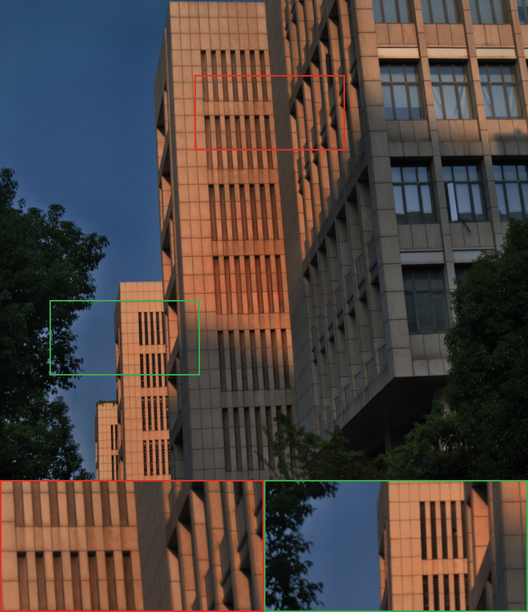} &
    \includegraphics[width=0.16\linewidth]{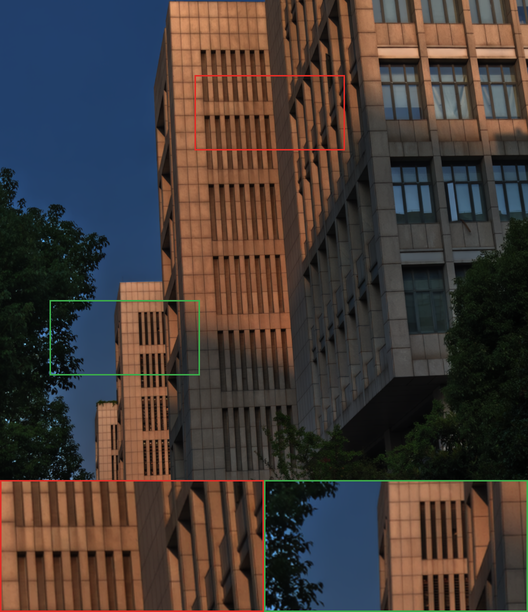} \\
    {\footnotesize LightenDiffusion} & {\footnotesize DnLUT} & {\footnotesize CanonCGT} &
    {\footnotesize ShiftLUT} & {\footnotesize LoopLUT (ours)} & {\footnotesize GT} \\
  \end{tabular}
  \caption{Qualitative comparison on the Mobile-Spec test set. Layout as in
  Figure~\ref{fig:qual_lcdp}; the red crop covers the sunlit facade and the
  green crop the backlit region at lower left, examining color reproduction
  under warm illumination and tonal separation in the backlit building and tree
  canopy.}
  \label{fig:qual_mobile}
\end{figure}

Figures~\ref{fig:qual_lcdp}--\ref{fig:qual_mobile} show the same trend. On
LCDP, RUAS overexposes the white wall and HDRNet and AdaInt leave the steel frame
in shadow, whereas LoopLUT brightens the frame without clipping the wall. On
FiveK, HDRNet leaves a yellow-green cast and CSEC and LCDPNet fail visibly,
while LoopLUT matches the target color temperature. On Mobile-Spec, RUAS turns
almost the whole image white and HDRNet and LCDPNet cannot lift the facade;
LoopLUT brightens the facade while keeping the sky suppressed.

\paragraph{Real-world photographs.} To examine performance in the real world
beyond standard test sets, we compare LoopLUT with CSEC and LCDPNet on two
everyday smartphone photographs: LoopLUT attains the best average NIQE and
BRISQUE while running 11$\times$ faster than LCDPNet and 99$\times$ faster than
CSEC (Figure~\ref{fig:realworld} in Appendix~\ref{sec:appendix_realworld}).

\subsection{Ablation studies}
\label{sec:ablation}

\begin{figure}[t]
  \centering
  \begin{minipage}[t]{0.54\linewidth}
    \vspace{0pt}
    \centering
    \includegraphics[width=\linewidth]{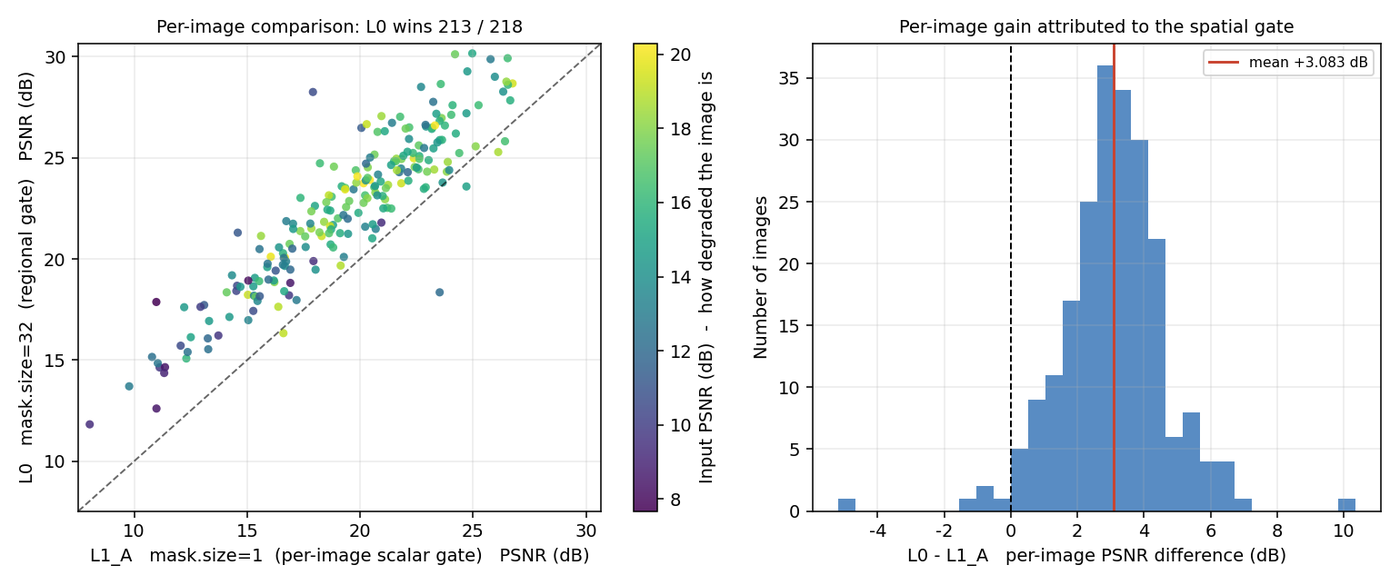}
    \captionof{figure}{Ablation~1: region gate ($s{=}32$) vs.\ per-image
    scalar gate ($s{=}1$), same rounds and parameters, 218 LCDP test images.
    Left: per-image PSNR. Right: per-image gain.}
    \label{fig:abl_spatial}
  \end{minipage}\hfill
  \begin{minipage}[t]{0.42\linewidth}
    \vspace{0pt}
    \centering
    \captionof{table}{Ablation~2: number of rounds $K$ (LCDP). Best in bold;
    $\dagger$: main configuration; ms: latency at $1620\times1080$.}
    \label{tab:ksweep}
    \vspace{2pt}
    \small
    \setlength{\tabcolsep}{2.5pt}
    \begin{tabular}{@{}lccccc@{}}
      \toprule
      $K$ & PSNR$\uparrow$ & SSIM$\uparrow$ & LPIPS$\downarrow$ & CIEDE$\downarrow$ & ms \\
      \midrule
      1 & 22.62 & 0.8214 & 0.1657 & 7.64 & 2.49 \\
      2 & 23.17 & 0.8428 & 0.1603 & 7.56 & 3.87 \\
      3$^{\dagger}$ & 23.48 & 0.8616 & 0.1558 & 7.53 & 5.10 \\
      4 & \textbf{23.61} & \textbf{0.8671} & \textbf{0.1539} & \textbf{7.49} & 6.32 \\
      5 & 23.60 & 0.8669 & 0.1541 & \textbf{7.49} & 7.49 \\
      6 & 23.56 & 0.8658 & 0.1545 & 7.50 & 8.71 \\
      7 & 23.51 & 0.8639 & 0.1552 & 7.51 & 9.83 \\
      8 & 23.43 & 0.8607 & 0.1566 & 7.54 & 11.34 \\
      \bottomrule
    \end{tabular}
  \end{minipage}
\end{figure}

\paragraph{Effectiveness of spatial gating.} To separate spatial adaptivity
from multi-round cascading, we keep the three-round cascade and the full
capacity but lower the gate bottleneck from $s{=}32$ to $s{=}1$, so every round
can only act on the whole image. Region gating wins on 213/218 test images with
a mean per-image PSNR gain of $+3.08$\,dB (Figure~\ref{fig:abl_spatial}): with
an identical cascade, using different LUTs in different regions accounts for
the bulk of the gain.

\paragraph{Effectiveness of the number of rounds.} We sweep $K{=}1$ to $8$ and
measure accuracy and latency together (Table~\ref{tab:ksweep}). Accuracy rises
with diminishing returns, the second, third and fourth rounds adding $+0.55$,
$+0.31$ and $+0.13$\,dB, peaks on all four metrics at $K{=}4$ (23.61\,dB) and
declines slowly from $K{\ge}5$, as the contracting coverage bound leaves later
rounds ever smaller regions to correct. Cost grows linearly, by
1.1--1.5\,ms per round at $1620\times1080$, so the gain per added millisecond
falls from 0.40\,dB/ms for the second round to 0.25 for the third and 0.11 for
the fourth: $K{=}4$ would buy $+0.13$\,dB with 24\% more latency. We adopt
$K{=}3$ as the last round whose gain is commensurate with its cost.

\begin{figure}[htbp]
  \centering
  \setlength{\tabcolsep}{1.5pt}
  \renewcommand{\arraystretch}{0.6}
  \begin{tabular}{cccccc}
    & {\footnotesize Current input} & {\footnotesize Gate $\mathbf{m}_k$}
      & {\footnotesize Hard selection} & {\footnotesize Partition overlay}
      & {\footnotesize Round output} \\
    \rotatebox{90}{\footnotesize ~~Round 1}
      & \includegraphics[width=0.155\linewidth]{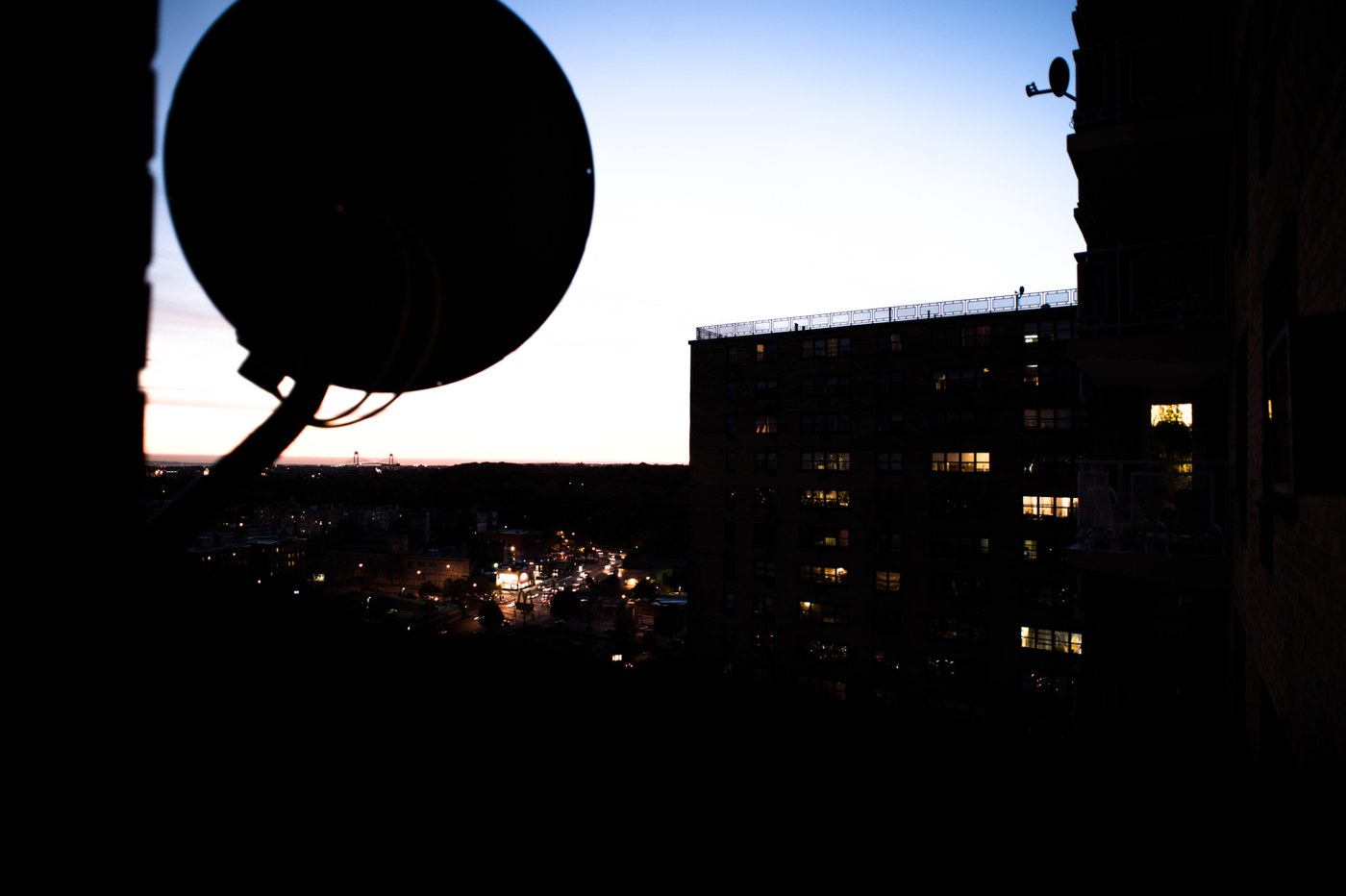}
      & \nacell & \nacell %
      & \includegraphics[width=0.155\linewidth]{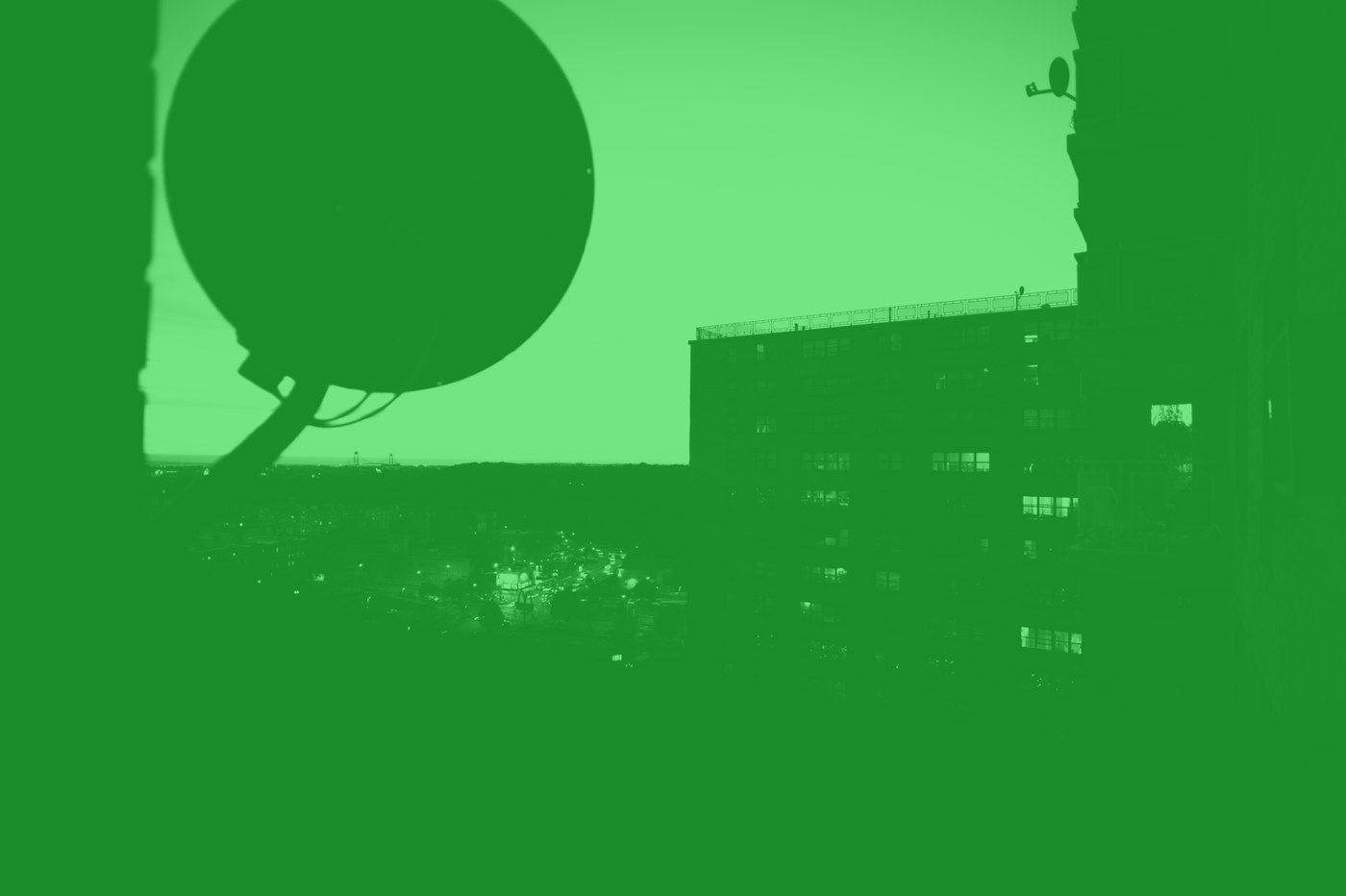}
      & \includegraphics[width=0.155\linewidth]{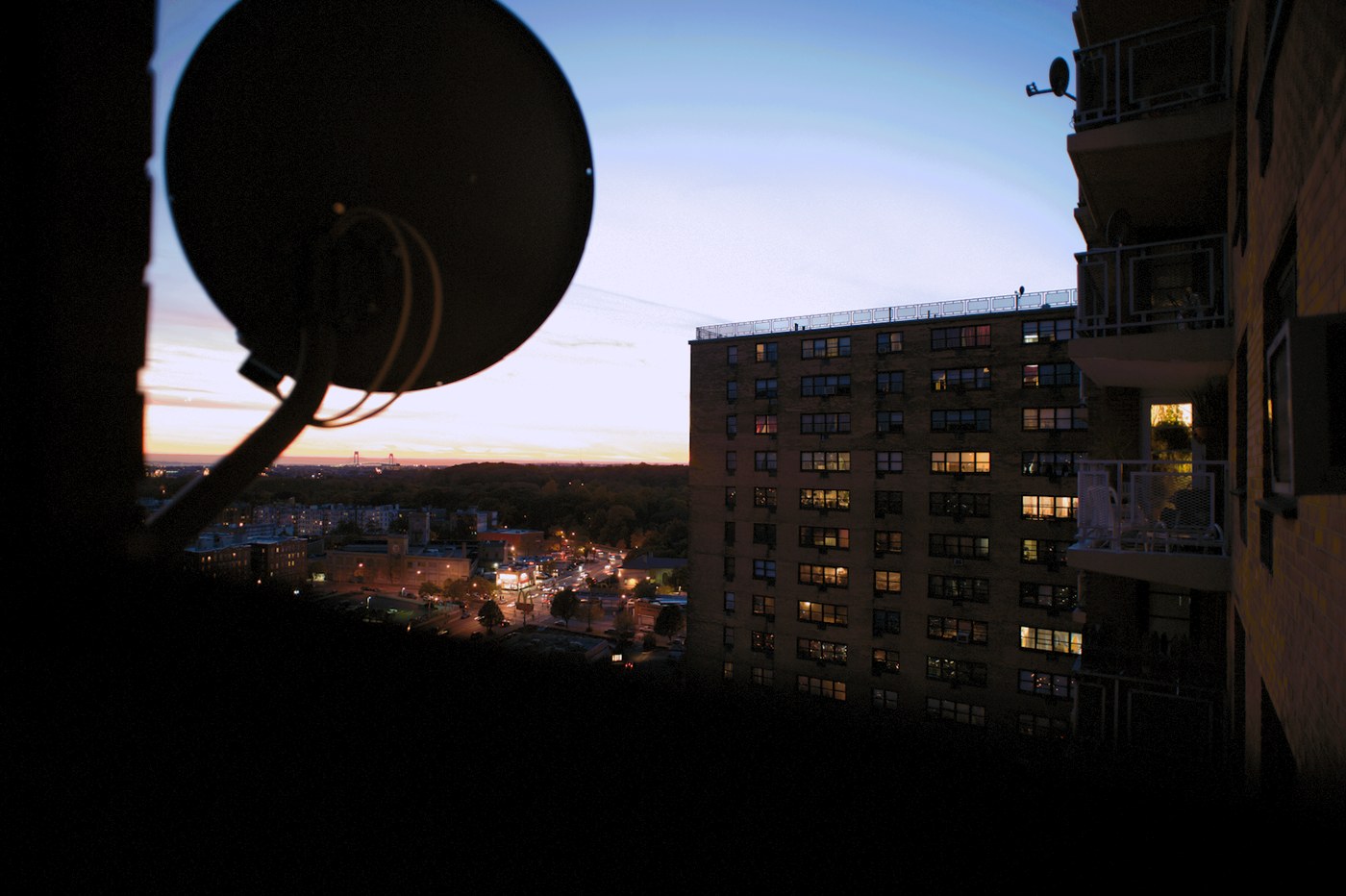} \\
    & {\scriptsize 19.16\,dB} & & & & {\scriptsize 20.56\,dB} \\[4pt]
    \rotatebox{90}{\footnotesize ~~Round 2}
      & \includegraphics[width=0.155\linewidth]{figures/abl3_r1_out.jpg}
      & \includegraphics[width=0.155\linewidth]{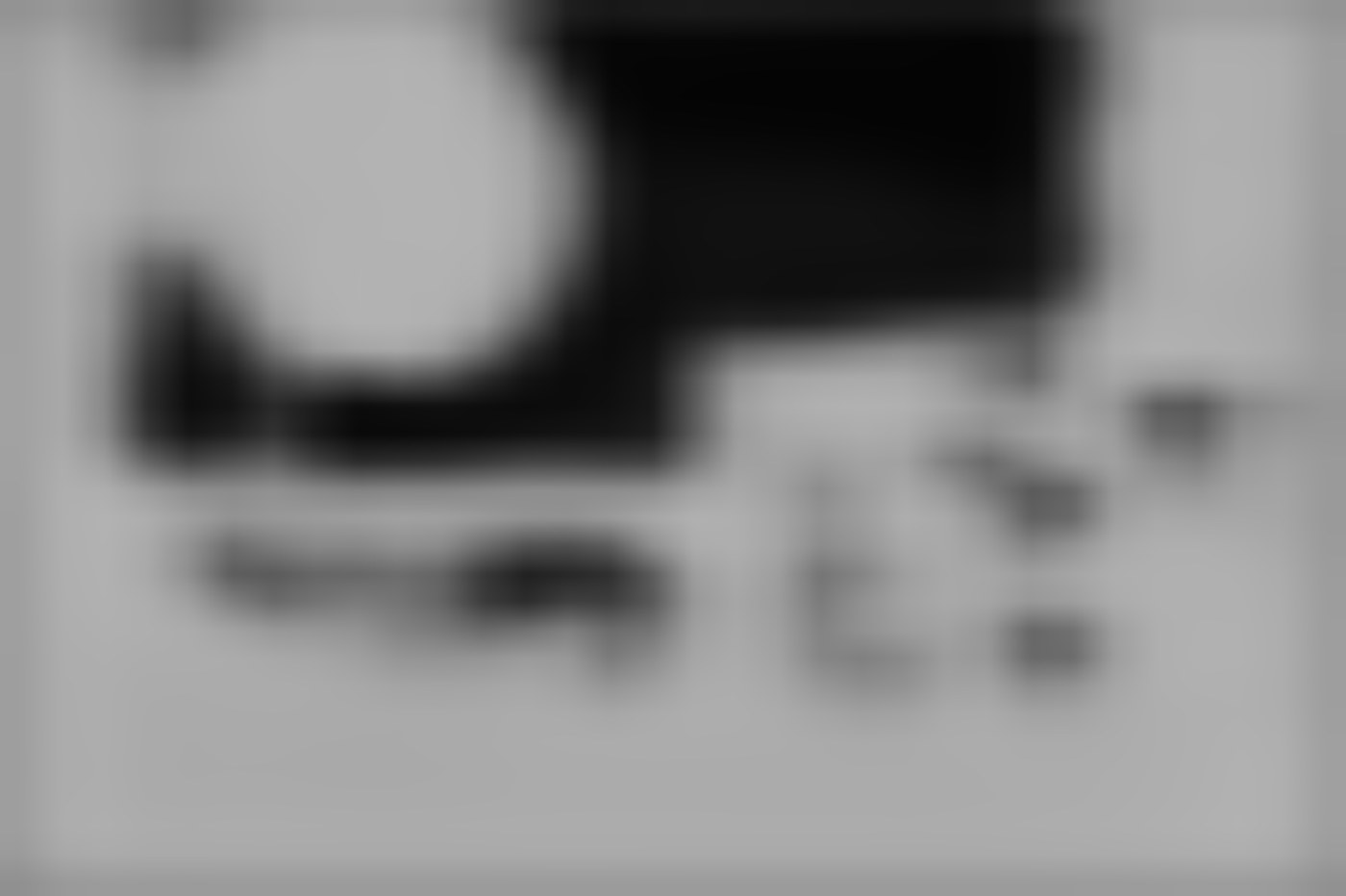}
      & \includegraphics[width=0.155\linewidth]{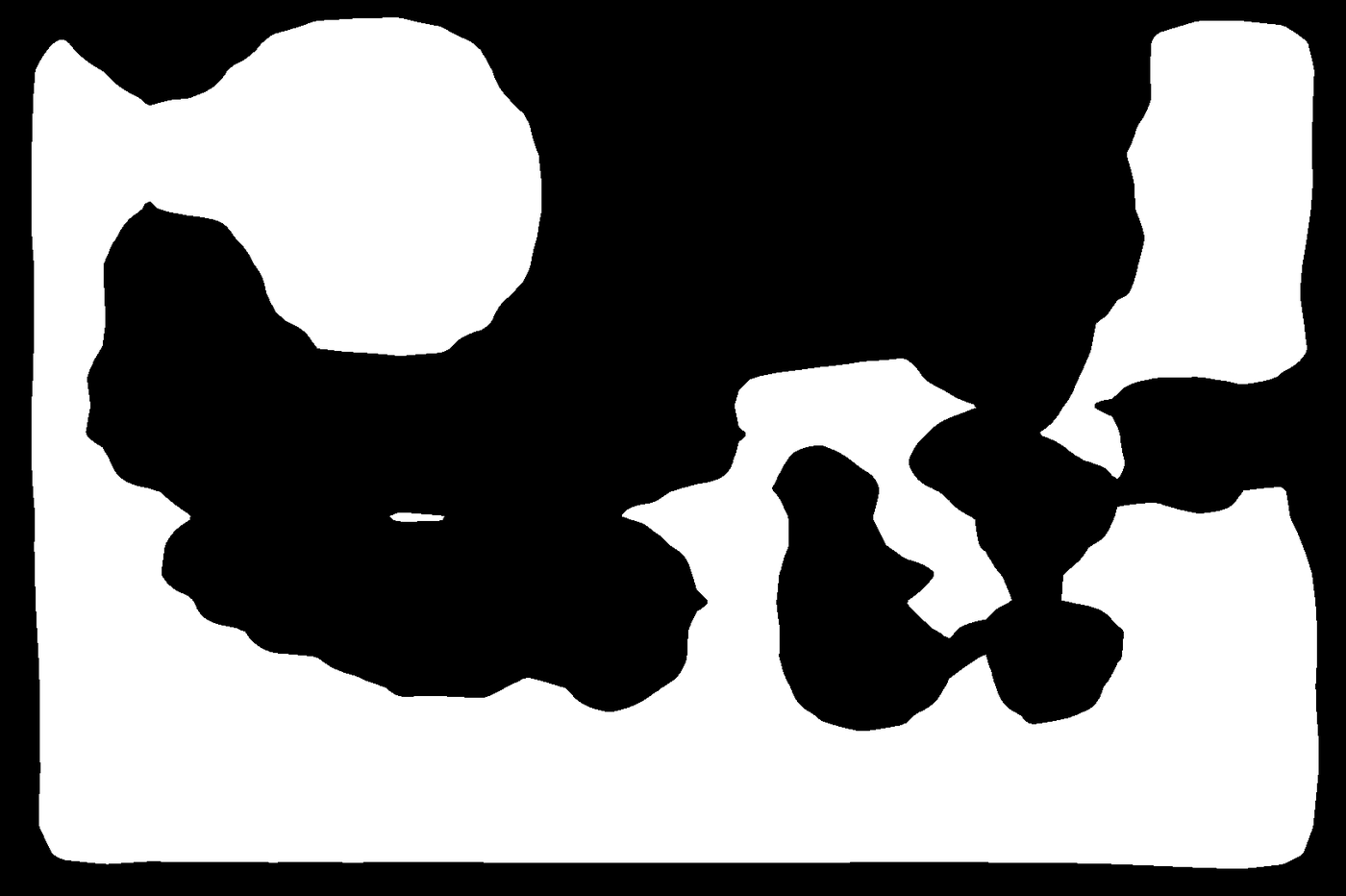}
      & \includegraphics[width=0.155\linewidth]{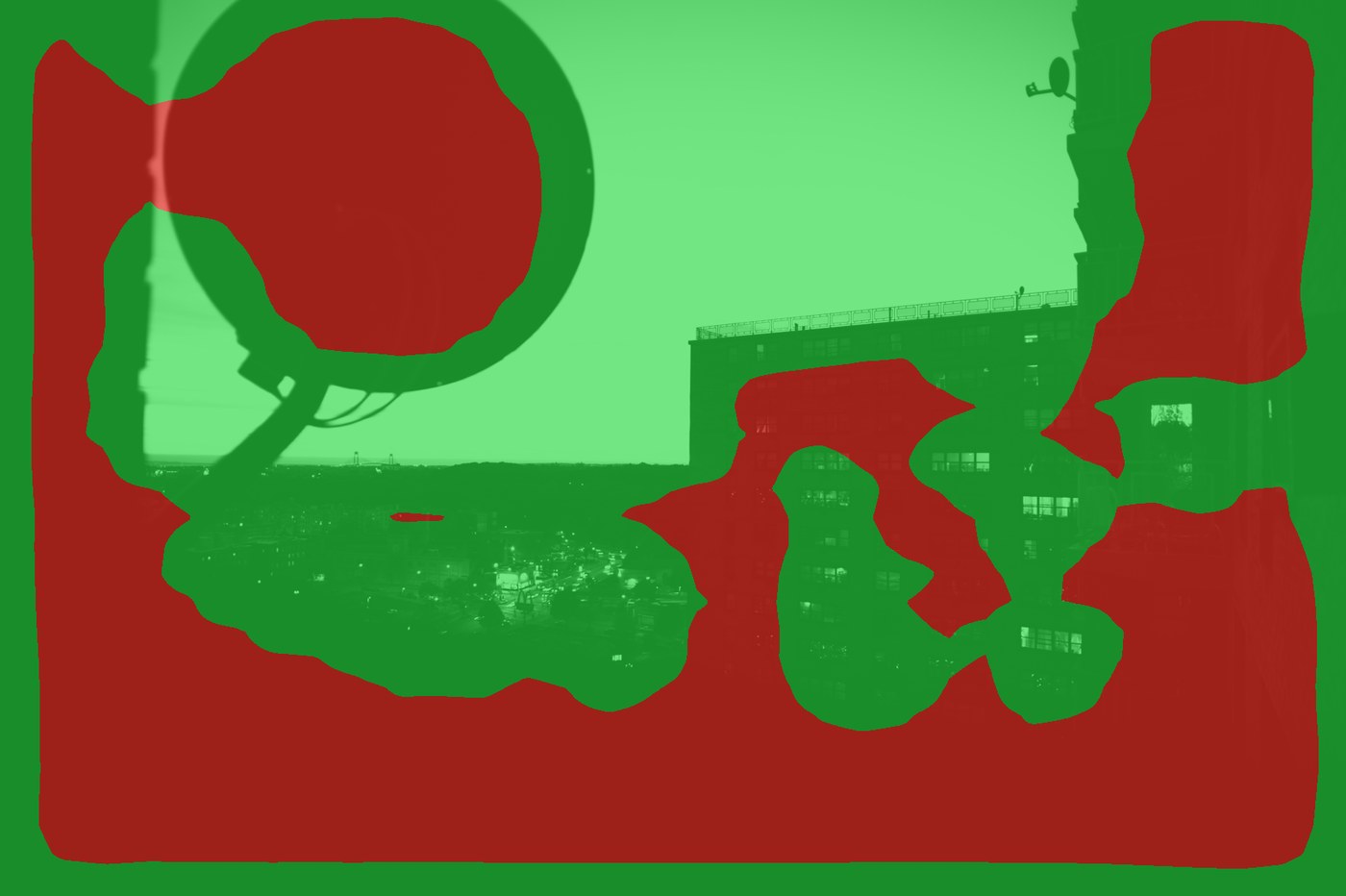}
      & \includegraphics[width=0.155\linewidth]{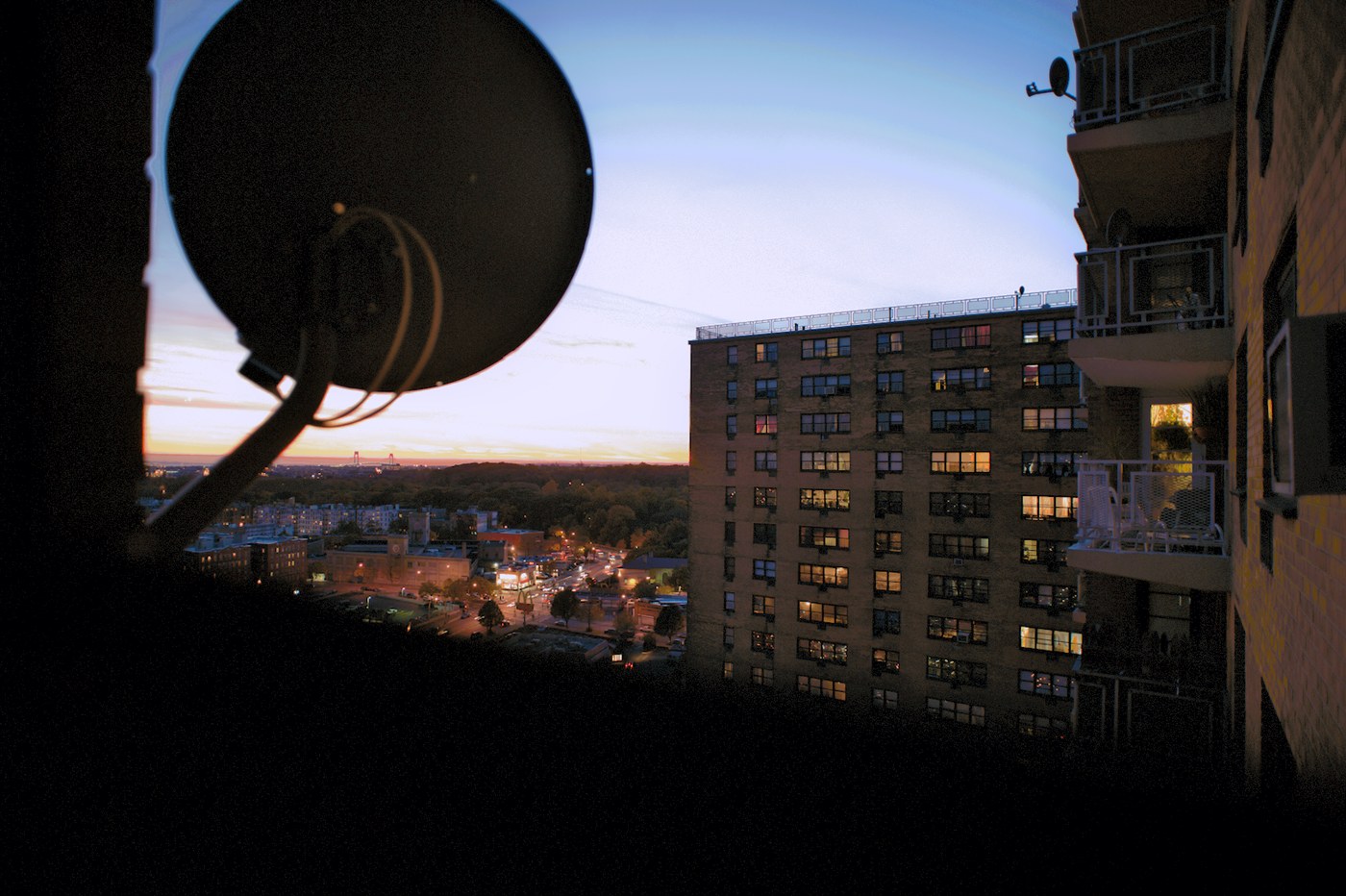} \\
    & {\scriptsize 20.56\,dB} & & & & {\scriptsize 23.74\,dB} \\[4pt]
    \rotatebox{90}{\footnotesize ~~Round 3}
      & \includegraphics[width=0.155\linewidth]{figures/abl3_r2_out.jpg}
      & \includegraphics[width=0.155\linewidth]{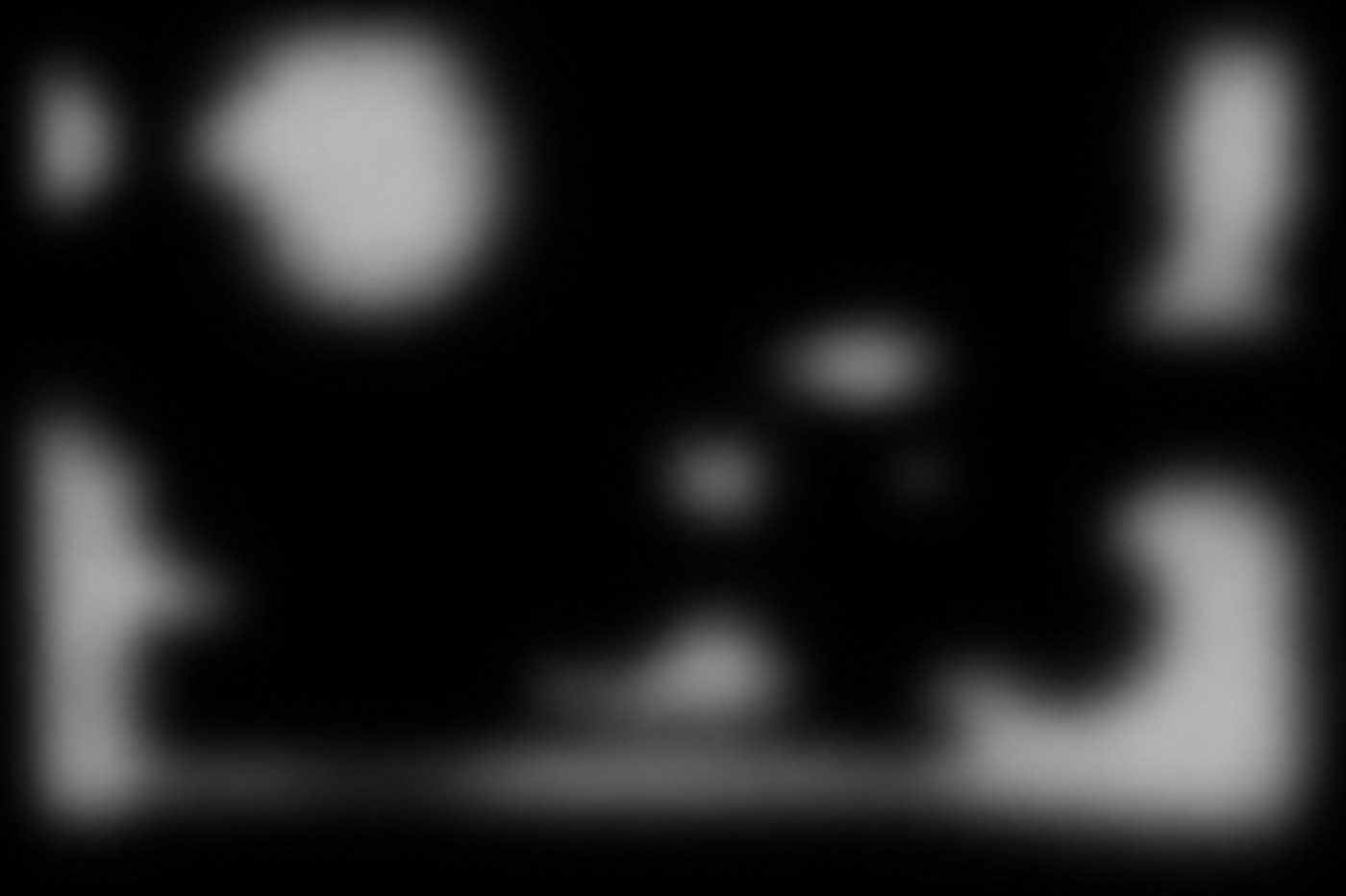}
      & \includegraphics[width=0.155\linewidth]{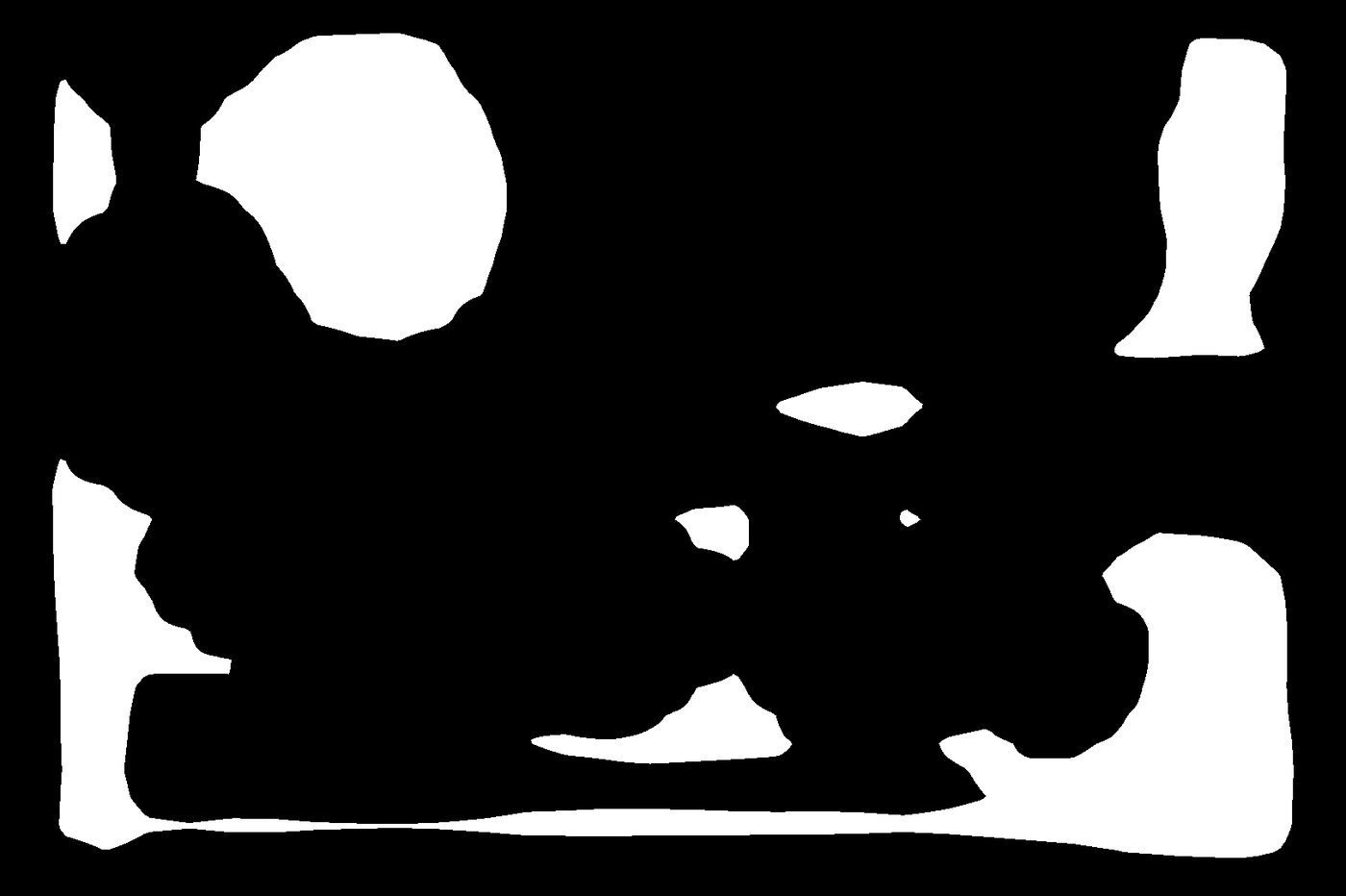}
      & \includegraphics[width=0.155\linewidth]{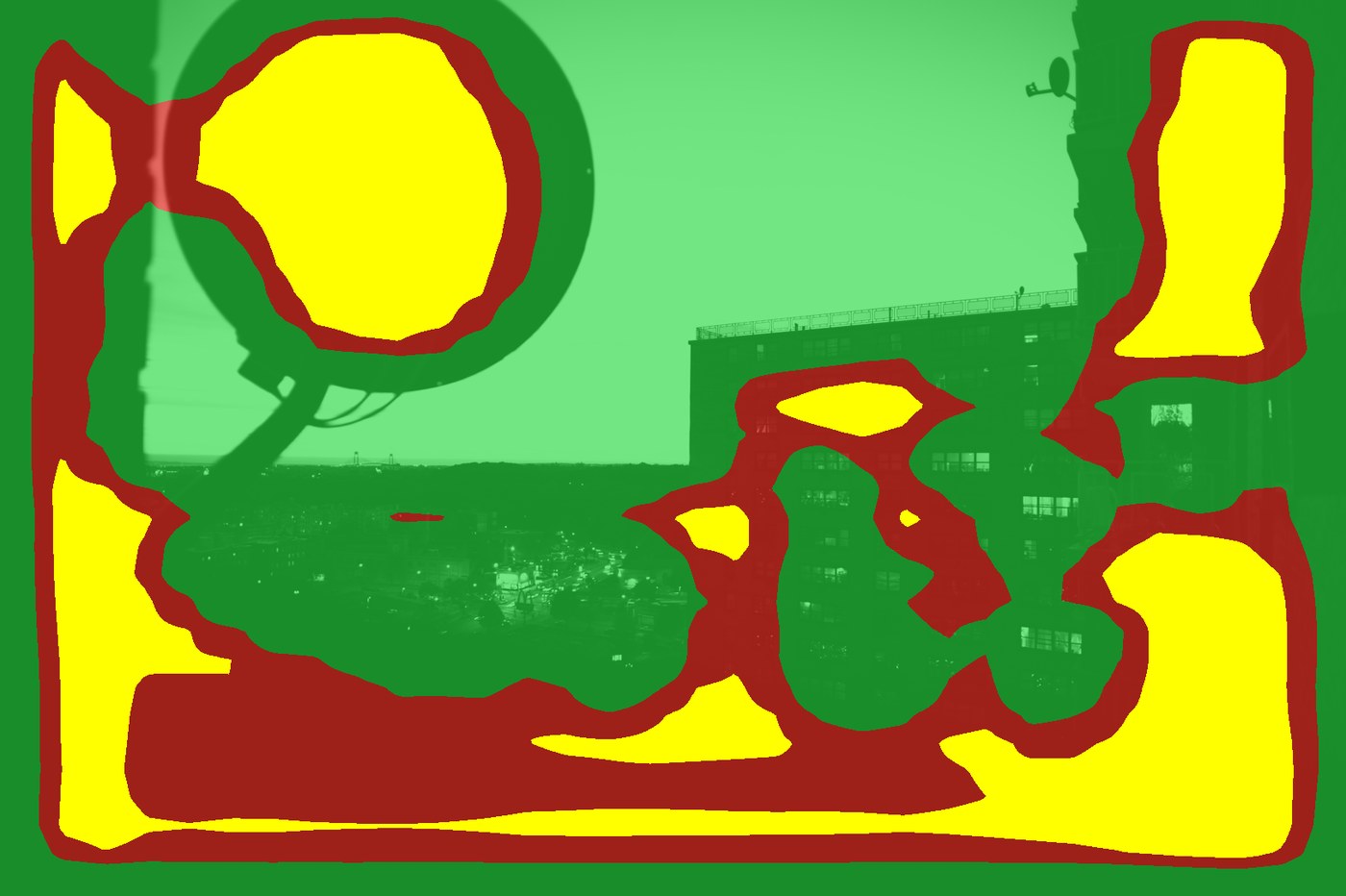}
      & \includegraphics[width=0.155\linewidth]{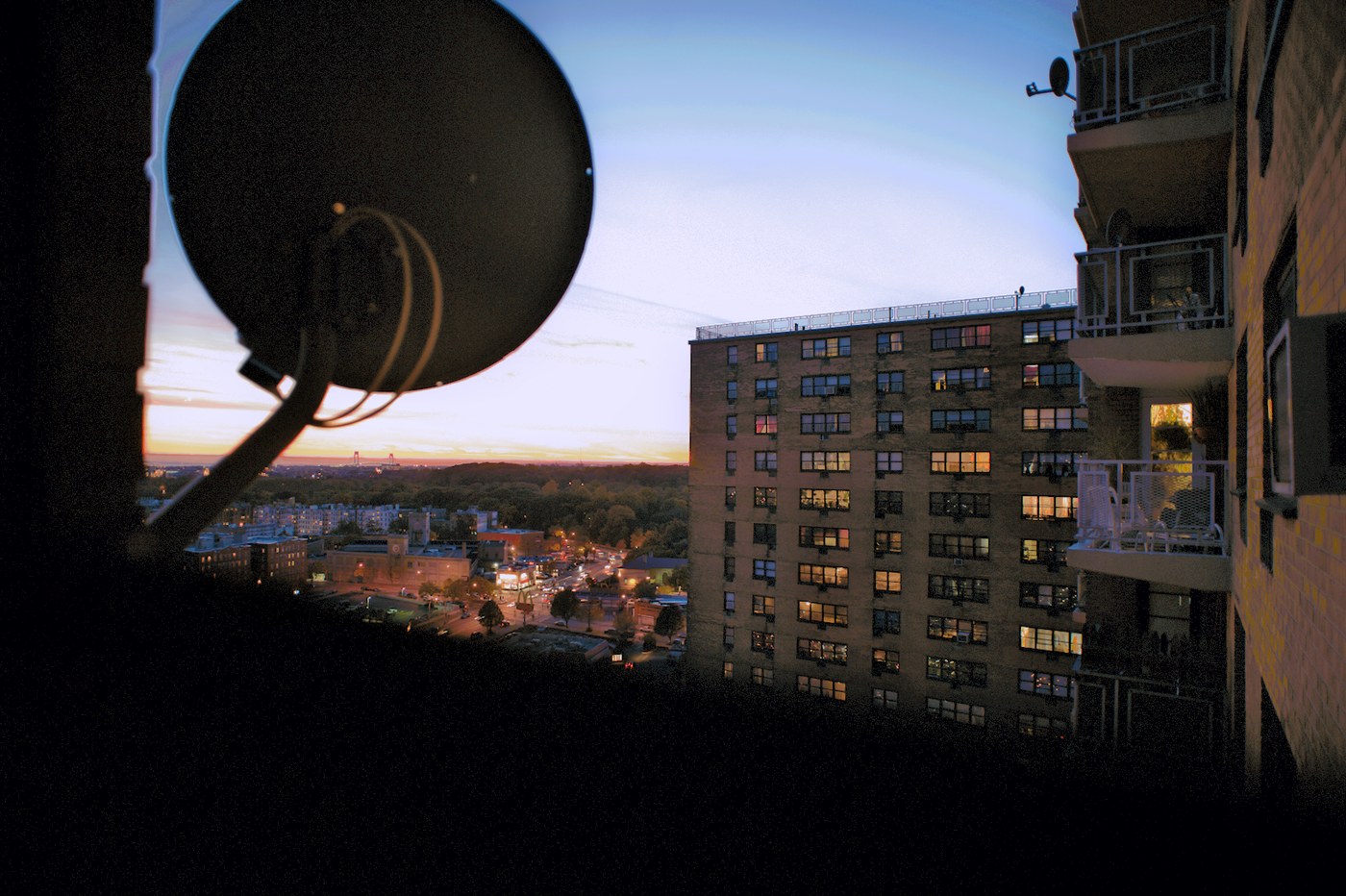} \\
    & {\scriptsize 23.74\,dB} & & & & {\scriptsize 24.04\,dB} \\
  \end{tabular}
  \caption{Ablation~3: progressive refinement on an LCDP test example. PSNR
  rises 19.16 (input) $\to$ 20.56 $\to$ 23.74 $\to$ 24.04\,dB, and each row's
  ``Current input'' is the previous row's ``Round output''. Round 1 is the
  global round, so its overlay is entirely green. In the partition overlay,
  green marks pixels fixed after round 1 ($\mathbf{A}_1$), red those entering round 2
  ($\mathbf{A}_2$) and yellow round 3 ($\mathbf{A}_3$); ``Hard selection'' is a thresholded view
  of the gate, shown for illustration only.}
  \label{fig:progressive}
\end{figure}

\paragraph{Effectiveness of progressive refinement process.} Figure~\ref{fig:progressive}
unfolds the decision process of a test example round by round. The global round
performs the overall brightening ($+1.4$\,dB); the round-2 gate skips the
already correctly graded sky and admits the underexposed foreground and
building, contributing the bulk of the gain ($+3.2$\,dB); round 3 contracts to
the heaviest residual areas for a small refinement ($+0.3$\,dB). PSNR rises
monotonically while the corrected region contracts monotonically, consistent
with the support contraction of Equation~\eqref{eq:partition}, which indicates
that the gates allocate later capacity to positions the previous round failed to
handle rather than reapplying the same correction everywhere.

\subsection{Computational efficiency}
\label{sec:complexity}

\begin{table}[htbp]
  \centering
  \caption{Resolution scaling and memory behavior of LoopLUT. The
  decision stage is constant at 6.135~GFLOPs and 1.84~ms \emph{at every
  resolution}; only the pure-lookup execution stage grows with pixel count.
  Traffic is the
  per-frame device-memory read${+}$write volume of the current unfused
  implementation, and Bandwidth is that volume divided by the measured
  latency.}
  \label{tab:complexity}
  \small
  \setlength{\tabcolsep}{3pt}
  \resizebox{\columnwidth}{!}{%
  \begin{tabular}{@{}lrrrrrrrrrr@{}}
    \toprule
    Resolution & Pixels & \multicolumn{3}{c}{GFLOPs} & \multicolumn{3}{c}{Latency}
               & \multicolumn{3}{c}{Device memory} \\
    \cmidrule(lr){3-5}\cmidrule(lr){6-8}\cmidrule(lr){9-11}
     & (MP) & Decide & Exec & Total & Decide & Total & Through- & Peak & Traffic & Band- \\
     &      &        &      &       & (ms)   & (ms)  & put (FPS) & (MB) & (MB) & width (GB/s) \\
    \midrule
    $1620\times 1080$ (LCDP native) & 1.75 & 6.135 & 0.362 & 6.497 & 1.84 & 5.10  & 196 & 93  & 774  & 159 \\
    $1920\times 1080$ (FHD)         & 2.07 & 6.135 & 0.429 & 6.564 & 1.84 & 5.67  & 176 & 104 & 880  & 163 \\
    $3840\times 2160$ (4K)          & 8.29 & 6.135 & 1.717 & 7.852 & 1.84 & 16.23 & 62  & 296 & 2914 & 188 \\
    \bottomrule
  \end{tabular}}
\end{table}

LoopLUT has 2.89\,M parameters (0.54\,M of them basis LUTs), fp32 weights of
11.04\,MB (5.53\,MB in fp16), and needs 6.564\,GFLOPs to process an FHD frame,
which it does in 5.67\,ms (176\,FPS).

\paragraph{Why the cost barely grows with resolution.}
The decision stage (one backbone pass, $K{-}1$ gating heads and $K$ MLP
evaluations) runs at a constant $256\times 256$, costing 6.135\,GFLOPs and
1.84\,ms at any output resolution; the execution stage is $K$ trilinear lookups
plus a per-pixel multiply-accumulate, about 69\,FLOPs per pixel per round. From
$1620\times 1080$ to 4K the pixel count grows $4.7\times$ but total computation
only 21\% (6.497 to 7.852\,GFLOPs), an asymptotic advantage at ultra-high
resolution over spatially aware methods whose decision cost grows with it.

\paragraph{FLOPs are not the whole story.}
Latency does not follow that 21\%: it triples, from 5.10 to 16.23\,ms. Table~\ref{tab:complexity}
shows why. The decision stage is 94\% of the FLOPs at $1620\times 1080$ but only
36\% of the time, and at 4K it is still 78\% of the FLOPs yet just 11\% of the
time. What grows is the execution stage, and it is bound by memory rather than
arithmetic: at 4K the current implementation moves 2.9\,GB per frame, or
188\,GB/s. The
execution stage is therefore running close to the hardware's memory limit, not
leaving a design inefficiency on the table.

This also bounds what an optimized implementation can gain. The 2.9\,GB figure
is the traffic of an unfused pipeline in which every operator round-trips
through device memory; a single fused kernel need only read the input image and
write the output, 94.9\,MB at 4K in fp16, a $31\times$ reduction.

\FloatBarrier

\subsection{Discussion}
\label{sec:general}
\label{sec:discussion}

\begin{wraptable}{r}{0.45\linewidth}
  \vspace{-12pt}
  \centering
  \caption{LSUI underwater benchmark. Best in bold, second underlined;
  compared methods quoted from public reports.}
  \label{tab:lsui}
  \footnotesize
  \setlength{\tabcolsep}{2.5pt}
  \begin{tabular}{@{}lccc@{}}
    \toprule
    Method & Venue & PSNR$\uparrow$ & SSIM$\uparrow$ \\
    \midrule
    UGAN & ICRA'18 & 19.29 & 0.7943 \\
    UIE-DAL & CVPRW'19 & 16.57 & 0.75 \\
    Ucolor & TIP'21 & 21.45 & 0.83 \\
    U-shape Trans. & TIP'23 & \underline{25.13} & \underline{0.86} \\
    \midrule
    LoopLUT (ours) & -- & \textbf{25.37} & \textbf{0.87} \\
    \bottomrule
  \end{tabular}
  \vspace{6pt}
  \setlength{\tabcolsep}{0.5pt}
  \begin{tabular}{ccc}
    \includegraphics[width=0.325\linewidth]{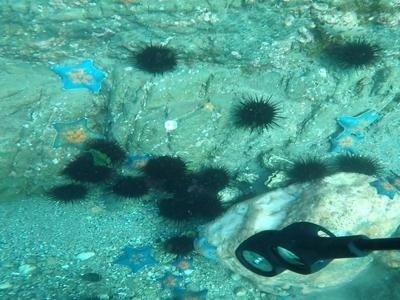} &
    \includegraphics[width=0.325\linewidth]{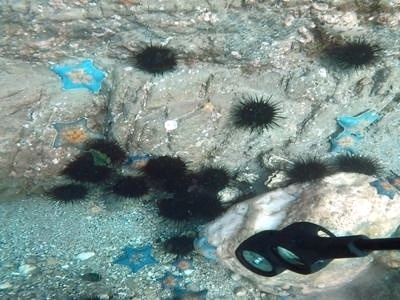} &
    \includegraphics[width=0.325\linewidth]{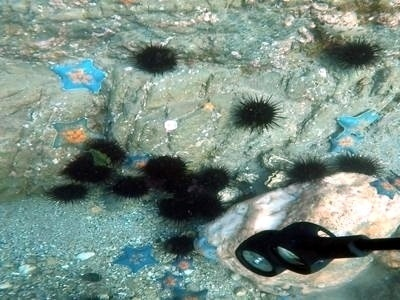} \\
    {\scriptsize Input} & {\scriptsize LoopLUT} & {\scriptsize GT} \\[-1pt]
    {\scriptsize 17.94\,dB\,/\,0.9045} & {\scriptsize 30.11\,dB\,/\,0.9640} & \\
  \end{tabular}
  \captionof{figure}{An LSUI test example; values under the panels are
  PSNR\,/\,SSIM.}
  \label{fig:lsui_main}
  \vspace{-10pt}
\end{wraptable}

\paragraph{Generality beyond exposure.} On LSUI the color shift comes from
medium absorption rather than exposure. With no underwater prior and unchanged
structural hyperparameters, LoopLUT reaches 25.37\,dB / 0.87,
above the task-specific U-shape Transformer (25.13\,dB / 0.86,
Table~\ref{tab:lsui}). It removes the cast without damaging structure
(Figure~\ref{fig:lsui_main}): the global round absorbs the whole-image cast and
the region rounds the depth-varying residual.
Appendix~\ref{sec:appendix_lsui} shows all methods (Figure~\ref{fig:lsui}).

\paragraph{What the loop contributes.} A single 3D LUT is a function of color
alone~\citep{zeng2020learning}, so two pixels with the same value receive the
same correction wherever they lie. The loop lifts this restriction yet keeps a
per-pixel lookup: each round adds a LUT restricted to the region its gate
selects, so identical colors in different regions are mapped differently
(Equation~\eqref{eq:partition}). As in recurrent
refinement~\citep{teed2020raft}, each gate reads the current enhancement state
and selects what is still wrong. Locality is what makes this work: a per-image
scalar gate with the same rounds and parameters falls $3.08$\,dB behind
(Figure~\ref{fig:abl_spatial}). The loop is also cheap, since all rounds share
one $256\times256$ backbone pass and each extra round adds 1.1--1.5\,ms
(Table~\ref{tab:ksweep}); as the coverage bound contracts, the benefit
concentrates in the early rounds, so $K{=}3$ suffices.

\paragraph{On-device deployment.} LoopLUT runs on an Android phone: we export
it to ONNX and execute it with the ONNX Runtime CPU backend, using the same fp32
weights as Table~\ref{tab:main} without quantization or pruning. The
application processes a photograph in roughly 100--200\,ms on CPU alone and
displays the input, the enhanced output and the partition assignment
(Appendix~\ref{sec:mobile}).

\FloatBarrier

\section{Conclusion}
\label{sec:conclusion}

We have presented LoopLUT, which performs iterative refinement through
low-resolution LUT decisions rather than repeated full-resolution image updates. Cascaded gates
organize $K$ LUTs, all applied directly to the original image, into a partition
of unity over that image, and fusion is carried out by the gates themselves, so
that no separate fusion module exists for a trivial solution to bypass and no
interpolation error is compounded through serial composition; the entire
decision process runs at a constant $256\times 256$, independent of the output
resolution. The accuracy of 23.48\,dB on LCDP, the best value on all four
metrics on FiveK and Mobile-Spec, and the cross-task generality check on LSUI
together indicate that spatial adaptivity, progressive refinement and a decision
cost independent of resolution can hold simultaneously within a single
architecture.

\section*{AI Use Statement}
Generative AI tools were used under the authors' direction to assist with
refining the research hypotheses, designing the research methodology and
experiments and providing feedback on them, implementing the method,
translation, cleaning and reformatting datasets, interpreting results, editing
the manuscript for readability, and typesetting and organizing the manuscript.
All AI-assisted code, data processing, analyses and text were reviewed and
verified by the authors. The authors take full responsibility for the final
content of this work.

\section*{Reproducibility Statement}
Code, configuration files and data-processing scripts will be released with
the paper. Section~\ref{sec:impl} gives the structural hyperparameters and the
training setup. Appendix~\ref{sec:appendix_proof} proves the structural
properties of the cascaded partition, Appendix~\ref{sec:appendix_design} gives
the definition and initialization of every loss term, and
Appendix~\ref{sec:appendix_data} gives the dataset splits and the evaluation
protocol for all four benchmarks. All experimental results of our proposed
method, LoopLUT, reported in this paper can be reproduced with the released
code.

\bibliographystyle{iclr2027_conference}
\bibliography{refs}

\clearpage
\appendix
\makeatletter
\setlength{\@fptop}{0pt}
\setlength{\@fpsep}{12pt plus 2fil}
\setlength{\@fpbot}{0pt plus 1fil}
\makeatother
\startcontents[appendix]
\begin{center}
  {\LARGE\scshape Appendix}\par
  \vspace{6pt}
  {\color{citeblue}\rule{0.22\linewidth}{0.8pt}}\par
  \vspace{4pt}
  {\small\itshape Contents}
\end{center}
\vspace{6pt}
{\hypersetup{linkcolor=black}%
 \printcontents[appendix]{}{1}{\setcounter{tocdepth}{2}\setlength{\parskip}{0pt}}}
\clearpage

\section{Proofs of the structural properties of the cascaded partition}
\label{sec:appendix_proof}

This appendix proves the three structural properties of
Section~\ref{sec:cascade}. Only two facts are used throughout: the gates are
sigmoid outputs, so $\mathbf{m}_k(x)\in(0,1)$ pointwise; and the telescoping structure of
Equation~\eqref{eq:partition}.

\subsection{Preliminary: the tail-sum identity}
From $\mathbf{C}_{k+1}=\mathbf{C}_k\,\mathbf{m}_{k+1}$, the
region-round weights in Equation~\eqref{eq:partition} can be written in
difference form as $\mathbf{A}_k=\mathbf{C}_k(1-\mathbf{m}_{k+1})=\mathbf{C}_k-\mathbf{C}_{k+1}$ for $1<k<K$. Hence for any
$2\le j\le K$,
\begin{equation}
  \sum_{k=j}^{K} \mathbf{A}_k
  \;=\; \sum_{k=j}^{K-1}\bigl(\mathbf{C}_k-\mathbf{C}_{k+1}\bigr) + \mathbf{C}_K
  \;=\; \mathbf{C}_j,
  \label{eq:tailsum}
\end{equation}
that is, the cumulative gate $\mathbf{C}_j$ is exactly the total weight allocated to
round $j$ and all subsequent rounds.

\subsection{Partition of unity}
Setting $j=2$ in Equation~\eqref{eq:tailsum}
gives $\sum_{k=2}^{K}\mathbf{A}_k=\mathbf{C}_2$; together with $\mathbf{A}_1=1-\mathbf{m}_2=1-\mathbf{C}_2$, adding the two
yields $\sum_{k=1}^{K}\mathbf{A}_k\equiv 1$ pointwise. Non-negativity follows directly
from $\mathbf{m}_k(x)\in(0,1)$: $\mathbf{C}_k$ is a product of numbers in $(0,1)$ and
$1-\mathbf{m}_{k+1}\in(0,1)$, so $\mathbf{A}_k(x)>0$ for every $k$.

\subsection{Monotone contraction of the supports}
From $\mathbf{C}_{k+1}=\mathbf{C}_k\,\mathbf{m}_{k+1}$
and $\mathbf{m}_{k+1}\le 1$ we obtain the pointwise inequality $\mathbf{C}_{k+1}(x)\le \mathbf{C}_k(x)$, so
for any threshold $\varepsilon\ge 0$ the superlevel sets are nested,
$\{x:\mathbf{C}_{k+1}(x)>\varepsilon\}\subseteq\{x:\mathbf{C}_k(x)>\varepsilon\}$; taking
$\varepsilon=0$ gives $\supp(\mathbf{C}_{k+1})\subseteq\supp(\mathbf{C}_k)$. Combined with
$\mathbf{A}_K=\mathbf{C}_K$ and $\mathbf{A}_k=\mathbf{C}_k(1-\mathbf{m}_{k+1})$, which give $\supp(\mathbf{A}_K)=\supp(\mathbf{C}_K)$, the
chain $\supp(\mathbf{A}_K)\subseteq\supp(\mathbf{C}_K)\subseteq\cdots\subseteq\supp(\mathbf{C}_2)$ holds.
Since the nesting holds for every $\varepsilon$, the property does not depend on
reading ``support'' at the zero threshold: binarizing the soft gates at any
threshold yields a hard partition that contracts from round to round in the same
way.

\subsection{The per-pixel effective LUT}
Trilinear interpolation is linear in
the LUT grid values: for a fixed query color $\mathbf{c}$,
$\trilerp(\alpha \mathbf{T}+\beta \mathbf{S};\,\mathbf{c})=\alpha\,\trilerp(\mathbf{T};\mathbf{c})+\beta\,\trilerp(\mathbf{S};\mathbf{c})$.
Substituting the residual form $\mathbf{T}_k=\mathbf{T}_1+\sum_{j=2}^{k}\boldsymbol{\Delta}_j$ into
Equation~\eqref{eq:output} and exchanging the order of summation:
\begin{align}
  \hat{\mathbf{I}}(x)
  &= \sum_{k=1}^{K} \mathbf{A}_k(x)\,\mathbf{T}_k(\mathbf{I})(x)
   = \Bigl(\textstyle\sum_{k=1}^{K}\mathbf{A}_k(x)\Bigr)\,\mathbf{T}_1(\mathbf{I})(x)
   + \sum_{j=2}^{K}\Bigl(\textstyle\sum_{k=j}^{K}\mathbf{A}_k(x)\Bigr)\,\boldsymbol{\Delta}_j(\mathbf{I})(x)
   \notag\\
  &= \mathbf{T}_1(\mathbf{I})(x) + \sum_{j=2}^{K} \mathbf{C}_j(x)\,\boldsymbol{\Delta}_j(\mathbf{I})(x)
   \;=\; \mathbf{T}^{x}_{\text{eff}}(\mathbf{I})(x),
\end{align}
where the second equality uses the partition of unity and the tail-sum
identity~\eqref{eq:tailsum}; the last step uses linearity of interpolation to
merge, at each pixel, the linear combination of LUTs with fixed coefficients
$\{1,\mathbf{C}_j(x)\}$ back into a single effective LUT, which is
Equation~\eqref{eq:efflut}. \hfill$\square$

\section{Design rationale and training details}
\label{sec:appendix_design}

This appendix collects the design arguments and the loss definitions summarized
in Section~\ref{sec:cascade} to Section~\ref{sec:lutgen}.

\subsection{Why the gates must be the fusion}
A natural alternative is serial
composition, $\mathbf{I}_3=\mathbf{T}_3(\mathbf{T}_2(\mathbf{T}_1(\mathbf{I})))$, followed by a scalar or per-pixel weighted
fusion of the intermediate results $\{\mathbf{I}_1,\mathbf{I}_2,\mathbf{I}_3\}$. This alternative has two
structural defects. First, $\mathbf{I}_3$ already contains the full effect of the
preceding two rounds, so fusion merely selects among nested versions and its
optimum trivially collapses to picking the most complete one: we measure this
fusion to give only $0.005$\,dB over outputting $\mathbf{I}_3$ directly, so the fusion
module does no substantive work. Second, folding the serial composition into a
single LUT for deployment introduces a second-order interpolation error between
lattice points (measured up to $9.5/255$, enough to produce visible banding in
flat gradients such as sky). Equation~\eqref{eq:output} contains no separate
fusion step: the partition weights are the gates themselves, given per pixel and
trained jointly, and cannot be bypassed by any trivial solution; and because all
$K$ LUTs act directly on the original image $\mathbf{I}$, the compounded interpolation
error is eliminated structurally.

\subsection{The low-resolution gate bottleneck}
The bottleneck $s$ of
Equation~\eqref{eq:gate} is the core structural constraint of the gating module
rather than an approximation for saving computation, and it serves three
purposes. (i)~\textbf{Preventing a shortcut.} If the gate were generated at high
resolution with per-pixel color among its inputs, the network would take a
shortcut in which $\mathbf{m}_k(x)$ degenerates into a function $f(\mathbf{I}(x))$ of the pixel's
own color; by Equation~\eqref{eq:efflut}, $\mathbf{T}_{\text{eff}}$ would then remain a
pure function of $\mathbf{I}(x)$, the whole model would collapse into a single more
expressive global LUT, and spatial adaptivity would be lost. The bottleneck
makes each gate cell cover at least $8\times 8$ pixels of the original image, so
it cannot track color pixel by pixel and can only make region-level judgments;
the inputs are downsampled by area averaging rather than bilinearly, precisely
to prevent high-frequency color information from leaking in through aliasing.
(ii)~\textbf{Preventing seams.} A gate that varies too rapidly means that
adjacent pixels of the same color are mapped to different results, which is the
classic cause of halos and seams in local tone mapping; generating the gate at
low resolution is itself a strong smoothing prior. (iii)~\textbf{Preserving
efficiency.} The gate cost is fixed at $s\times s$, which preserves the property
that the decision is independent of resolution. The price is that the spatial
precision of the gate is bounded by $s$, so thin structures such as hair or
railings cannot be delineated on their own; this is a boundary we accept
deliberately. The design also rests on an asymmetry: judging \emph{where}
correction is still needed (a scalar map) is far easier than regressing the
correction itself (three channels per pixel), and the latter is precisely the
task carried by the LUT residual.

\subsection{Why the residual must be driven by region statistics}
If
$\boldsymbol{\Delta}_k$ were still generated from whole-image features, the gate would only
constrain \textbf{where} the correction is applied without letting the generator
\textbf{know} what is to be corrected, and the result would be no more than a
global LUT cropped by the gate. Masked weighted pooling inside the cumulative
gate therefore supplies the statistics of the region each round actually faces,
and differentiation between rounds becomes intrinsic to the structure rather
than a consequence of an explicit diversity regularizer. Giving the region
feature as a residual on the global feature, with the last layer of $P$
zero-initialized, makes the training starting point equivalent to using $\mathbf{F}_{\text{g}}$
directly, so region-specific correction directions emerge gradually during
training.

\subsection{An initialization constraint from the product structure}
In
Equation~\eqref{eq:efflut} the output depends on the parameters through the
product $\mathbf{C}_k\cdot\boldsymbol{\Delta}_k$, and the pre-clip residual is itself a product of coefficients
and bases. If either factor of a product is initialized to zero, the gradient of
the other factor is simultaneously zeroed, and this interlock cannot resolve
itself during training. We therefore set the final-layer bias of the gating head
to 0 (so the initial gate is $\approx 0.5$), rather than adopting a conservative
negative bias; and for the basis mixing head we zero only the bias while keeping
the default weight initialization. Together these ensure that both factors on
every product path receive gradient from the first step, at the cost of a
starting point that departs from the exact identity mapping by a visually
imperceptible amount (about 0.045).

\subsection{The region coverage term}
The coverage term of
Section~\ref{sec:lutgen} is
\begin{equation}
  \mathcal{L}_{\text{area}} = \sum_{k=2}^{K}
  \max\bigl(0,\ \bar{\mathbf{C}}_k-\tau_k\bigr)
  + \max\bigl(0,\ \tau_{\min}-\bar{\mathbf{C}}_k\bigr),
  \label{eq:area}
\end{equation}
where the upper bound $\tau_k$ decreases across rounds ($\tau_2>\tau_3$). We
deliberately avoid the intuitive scheme of admitting a pixel to the next round
whenever its error exceeds a threshold, for two reasons. First, a large error
does not imply that the pixel is correctable: a LUT is a per-pixel color
mapping, so if two pixels with the same RGB value inside a region require
opposite corrections, no residual $\boldsymbol{\Delta}_k$, however large, can satisfy both,
and selecting regions by error would allocate the limited correction budget to
irreparable pixels. Under an \textbf{upper bound} on coverage, by contrast,
which pixels are worth admitting is ordered automatically by the gradient of the
main loss according to \textbf{marginal benefit}, which is exactly the ordering
criterion required. Second, an absolute error threshold drifts over the course
of training: errors are large across the whole image early on, saturating the
gate, and later it becomes hard to select any region at all.
Equation~\eqref{eq:area} is an upper bound rather than an equality constraint:
the corrected region of a simple image may be far smaller than the bound, and
when degradation is mild the local correction naturally approaches zero. The
lower bound $\tau_{\min}$ is a small constant that keeps the gradient path open:
by the product structure above, once the gate collapses to 0 it interlocks with
the residual and cannot recover, and $\tau_{\min}$ pushes it back before the
collapse occurs.

\subsection{The smoothness and gate-supervision terms}
\textbf{Gate smoothness}
$\mathcal{L}_{\text{tv}}$ is the total variation of each $\mathbf{m}_k$, which together
with the low-resolution bottleneck suppresses halos and seams at region
boundaries. \textbf{LUT smoothness} $\mathcal{L}_{\text{smooth}}$ applies
total variation and a monotonicity hinge to each $\mathbf{T}_k$: smoothness alone does
not suppress local color inversions in which a brighter input yields a darker
output, and since such inversions are the main cause of color banding, a
monotonicity constraint penalizes them directly. \textbf{Gate supervision}
$\mathcal{L}_{\text{gate}}$ provides an auxiliary signal for the gate from
ground truth available only during training: the binary map of the
top-$(1{-}\tau_k)$ \textbf{per-image quantile} of the previous round's residual
correction serves as an auxiliary target for $\mathbf{m}_k$, sharing the same
hyperparameter $\tau_k$ as the coverage term so that no additional
hyperparameter is introduced; a per-image quantile rather than an absolute
threshold is used for the same reason as in Equation~\eqref{eq:area}. This
supervision exists only during training; at inference the gate is predicted
independently by Equation~\eqref{eq:gate} and no error computation is involved.

\section{Experimental setup}
\label{sec:appendix_data}

This appendix gives the splits, the evaluation protocol and the choice of
compared methods summarized in Sections~\ref{sec:setup} and~\ref{sec:impl}.

\subsection{Splits}
\textbf{LCDP}~\citep{wang2022lcdp}, mixed-exposure scenes in
which overexposure and underexposure coexist, 1{,}415/100/218 for
train/validation/test. \textbf{MIT-Adobe FiveK}~\citep{bychkovsky2011fivek},
expert retouching of RAW photographs with ExpertC as the target,
4{,}176/250/497. \textbf{Mobile-Spec}~\citep{zhou2024mobilespec}, mobile-phone
RGB paired with a GaiaSky-mini2 hyperspectral camera, 140/20/40.
\textbf{LSUI}~\citep{peng2023ushape}, underwater degradation caused by medium
absorption rather than exposure, split by an explicit list into 3{,}423/428/428.

\subsection{Evaluation protocol}
All four metrics are computed at native
resolution, without cropping or rescaling. LPIPS~\citep{zhang2018lpips} uses the
AlexNet backbone, and the reported color error is CIEDE2000 throughout.

\subsection{Selection of compared methods}
\label{sec:appendix_baselines}

The compared methods in Table~\ref{tab:main} are chosen to cover the approach
families most relevant to LoopLUT. (i) HDRNet~\citep{gharbi2017bilateral} is the
classic real-time enhancement paradigm. (ii) RUAS~\citep{liu2021ruas} and the
diffusion model LightenDiffusion~\citep{jiang2024lightendiffusion} represent
low-light enhancement, which brightens underexposed images. (iii)
LCDPNet~\citep{wang2022lcdp} and CSEC~\citep{li2024csec} are designed
specifically for over- and under-exposure correction, the degradation that LCDP
targets. (iv) AdaInt~\citep{yang2022adaint}, DnLUT~\citep{yang2025dnlut} and
ShiftLUT~\citep{zeng2026shiftlut} are LUT-based methods, which share the
efficiency motivation of LoopLUT. (v) CanonCGT~\citep{ko2026canoncgt} is a
recent reference-based color grading method, which maps the input to a canonical
representation before grading. Together, these baselines compare LoopLUT both
with methods that share its LUT design and with methods tailored to the
exposure scenarios in our benchmarks.

\section{Additional results}
\label{sec:appendix_results}

This appendix gives results that complement Section~\ref{sec:exp}: the average
over the three benchmarks of Table~\ref{tab:main}, a visual comparison on
LSUI, and real-world smartphone photographs.

\subsection{Average over the three benchmarks}
\label{sec:appendix_avg}

\begin{figure}[!ht]
  \centering
  \includegraphics[width=0.8\linewidth]{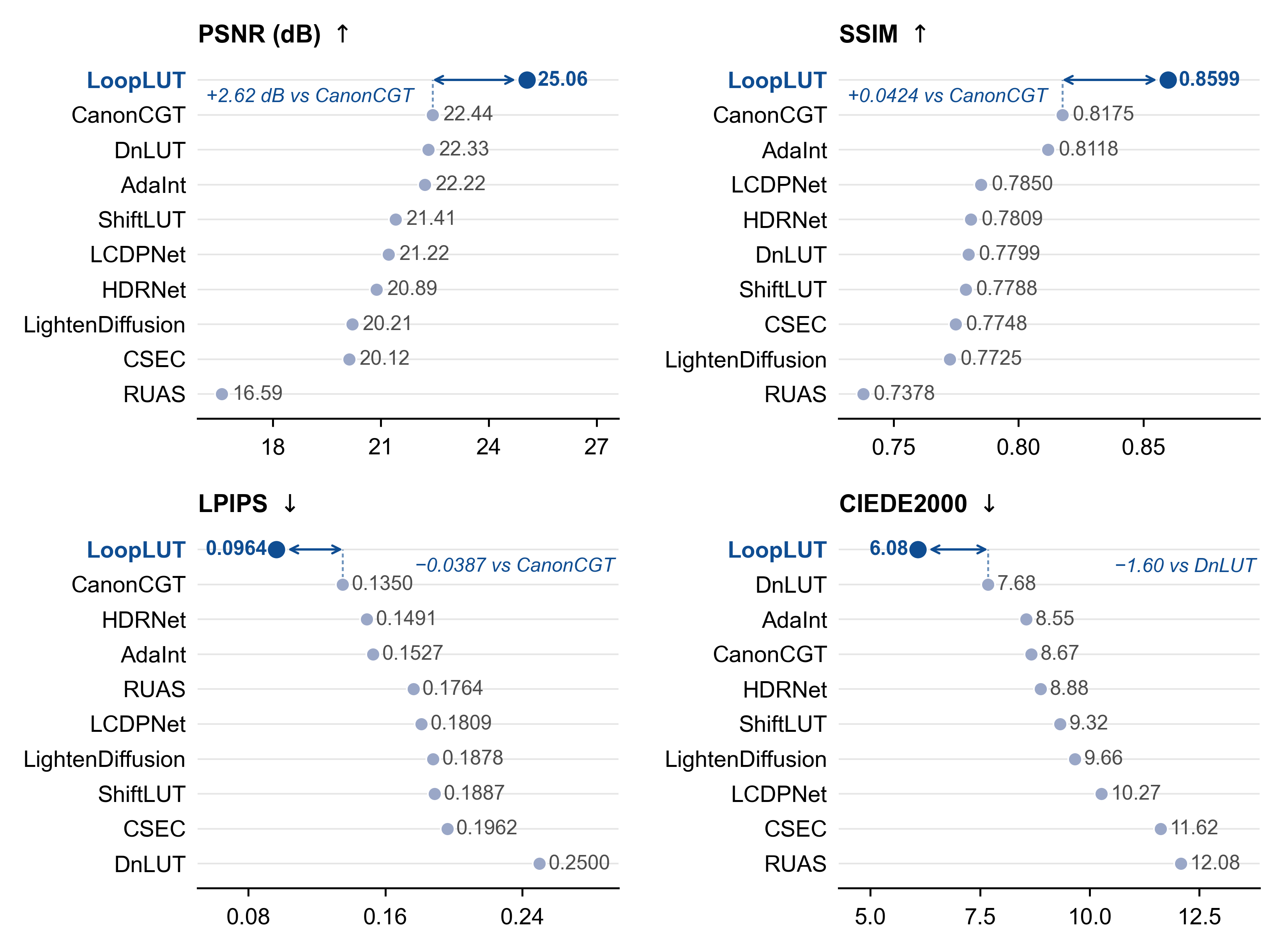}
  \caption{The averages of Table~\ref{tab:avg}, with methods sorted from best
  to worst in each panel; the arrow marks the gap between LoopLUT and the
  runner-up.}
  \label{fig:avg}
\end{figure}

Table~\ref{tab:avg} averages the results of each method in
Table~\ref{tab:main} over LCDP, MIT-Adobe FiveK and Mobile-Spec, with equal
weight per dataset, and Figure~\ref{fig:avg} plots the same averages per
metric. LSUI is excluded because its compared methods differ
(Table~\ref{tab:lsui}). LoopLUT ranks first on all four metrics. Its PSNR
(25.06\,dB) and SSIM (0.8599) exceed the runner-up CanonCGT by 2.62\,dB and
0.0424; its LPIPS (0.0964) is 0.0387 below CanonCGT, and its CIEDE2000 (6.08)
is 1.60 below the runner-up DnLUT.

\begin{table}[!ht]
  \centering
  \caption{Average over LCDP, MIT-Adobe FiveK and Mobile-Spec of the metrics in
  Table~\ref{tab:main}. In each column the \textbf{best} is bold and the
  \underline{second best} is underlined.}
  \label{tab:avg}
  \begin{tabular}{@{}lccccc@{}}
    \toprule
    Method & Venue & PSNR$\uparrow$ & SSIM$\uparrow$ & LPIPS$\downarrow$ & CIEDE$\downarrow$ \\
    \midrule
    HDRNet & SIGGRAPH'17 & 20.89 & 0.7809 & 0.1491 & 8.88 \\
    RUAS & CVPR'21 & 16.59 & 0.7378 & 0.1764 & 12.08 \\
    LCDPNet & ECCV'22 & 21.22 & 0.7850 & 0.1809 & 10.27 \\
    AdaInt & CVPR'22 & 22.22 & 0.8118 & 0.1527 & 8.55 \\
    LightenDiffusion & ECCV'24 & 20.21 & 0.7725 & 0.1878 & 9.66 \\
    CSEC & CVPR'24 & 20.12 & 0.7748 & 0.1962 & 11.62 \\
    DnLUT & CVPR'25 & 22.33 & 0.7799 & 0.2500 & \underline{7.68} \\
    CanonCGT & CVPR'26 & \underline{22.44} & \underline{0.8175} & \underline{0.1350} & 8.67 \\
    ShiftLUT & CVPR'26 & 21.41 & 0.7788 & 0.1887 & 9.32 \\
    \midrule
    LoopLUT (ours) & -- & \textbf{25.06} & \textbf{0.8599} & \textbf{0.0964} & \textbf{6.08} \\
    \bottomrule
  \end{tabular}
\end{table}

\subsection{Visual comparison on LSUI}
\label{sec:appendix_lsui}

\begin{figure}[!ht]
  \centering
  \setlength{\tabcolsep}{0.5pt}
  \renewcommand{\arraystretch}{0.5}
  \begin{tabular}{ccccccc}
    \includegraphics[width=0.138\linewidth]{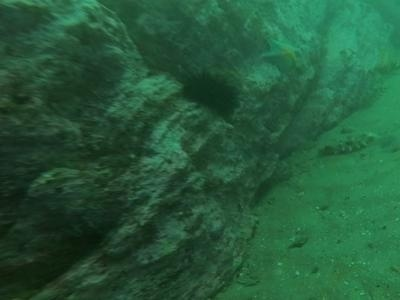} &
    \includegraphics[width=0.138\linewidth]{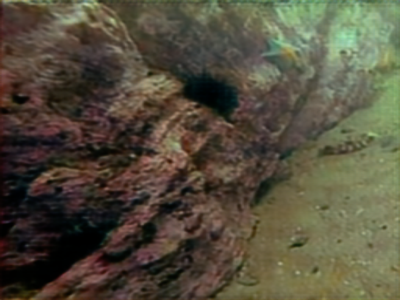} &
    \includegraphics[width=0.138\linewidth]{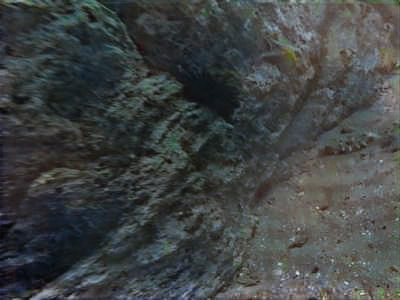} &
    \includegraphics[width=0.138\linewidth]{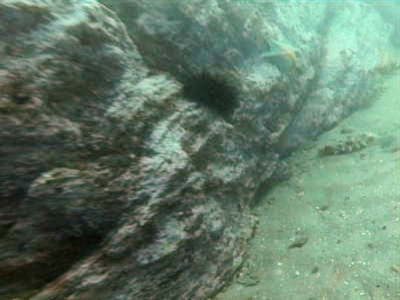} &
    \includegraphics[width=0.138\linewidth]{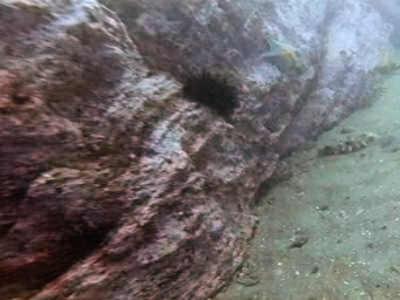} &
    \includegraphics[width=0.138\linewidth]{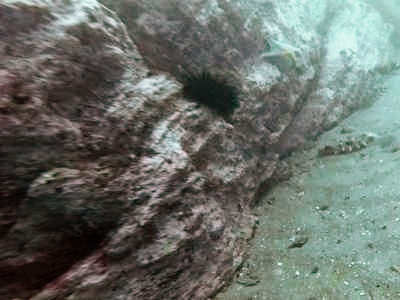} &
    \includegraphics[width=0.138\linewidth]{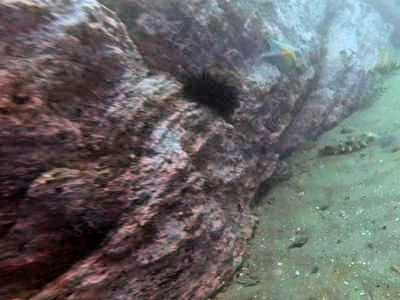} \\
    \includegraphics[width=0.138\linewidth]{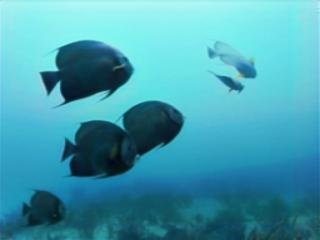} &
    \includegraphics[width=0.138\linewidth]{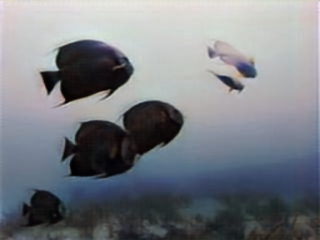} &
    \includegraphics[width=0.138\linewidth]{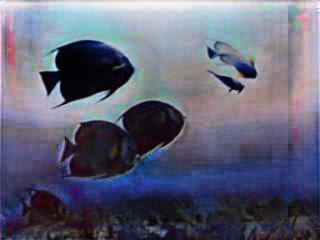} &
    \includegraphics[width=0.138\linewidth]{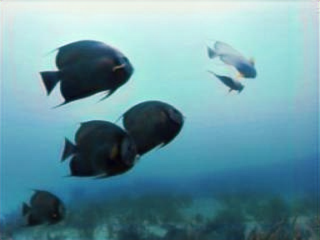} &
    \includegraphics[width=0.138\linewidth]{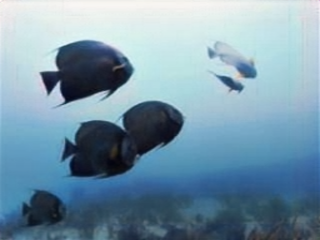} &
    \includegraphics[width=0.138\linewidth]{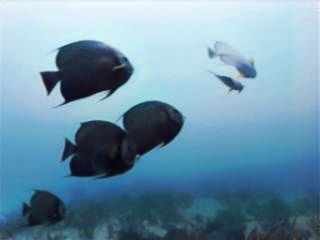} &
    \includegraphics[width=0.138\linewidth]{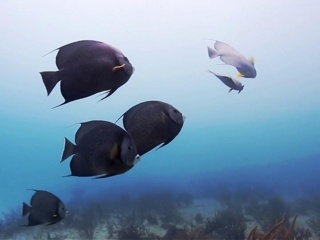} \\
    \includegraphics[width=0.138\linewidth]{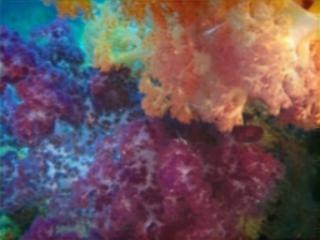} &
    \includegraphics[width=0.138\linewidth]{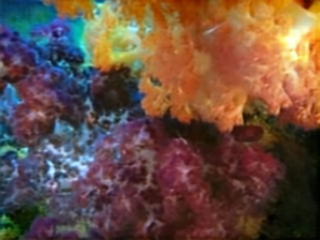} &
    \includegraphics[width=0.138\linewidth]{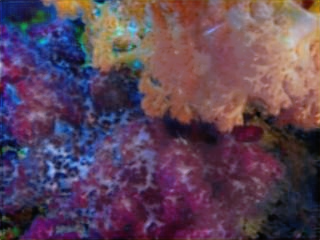} &
    \includegraphics[width=0.138\linewidth]{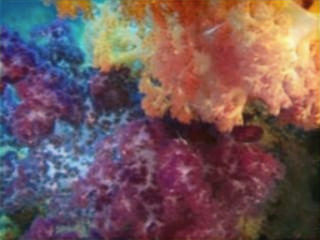} &
    \includegraphics[width=0.138\linewidth]{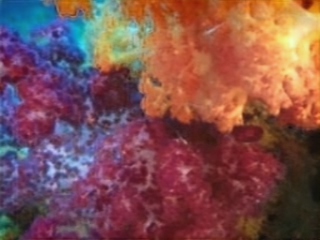} &
    \includegraphics[width=0.138\linewidth]{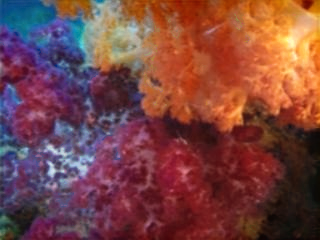} &
    \includegraphics[width=0.138\linewidth]{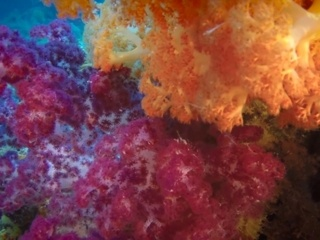} \\
    \includegraphics[width=0.138\linewidth]{figures/lsui_img4_input.png} &
    \includegraphics[width=0.138\linewidth]{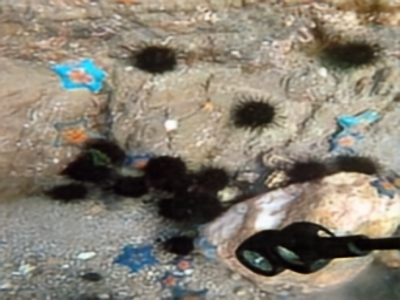} &
    \includegraphics[width=0.138\linewidth]{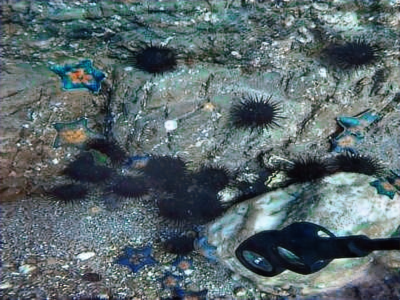} &
    \includegraphics[width=0.138\linewidth]{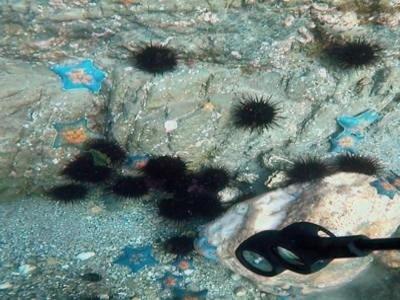} &
    \includegraphics[width=0.138\linewidth]{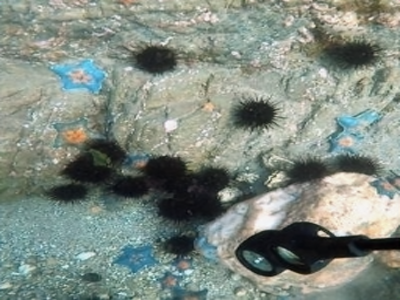} &
    \includegraphics[width=0.138\linewidth]{figures/lsui_img4_looplut.png} &
    \includegraphics[width=0.138\linewidth]{figures/lsui_img4_gt.png} \\
    {\footnotesize Input} & {\footnotesize UGAN} & {\footnotesize UIE-DAL} &
    {\footnotesize Ucolor} & {\footnotesize U-shape Trans.} &
    {\footnotesize LoopLUT} & {\footnotesize GT} \\
  \end{tabular}
  \caption{Visual comparison on four LSUI test images, one image per row. The
  compared methods are those of Table~\ref{tab:lsui}.}
  \label{fig:lsui}
\end{figure}

Figure~\ref{fig:lsui} compares LoopLUT
with the underwater enhancement methods of Table~\ref{tab:lsui} on four LSUI
test images, complementing the generality check of Section~\ref{sec:discussion}.

\subsection{Real-world photographs}
\label{sec:appendix_realworld}

\begin{figure}[!ht]
  \centering
  \setlength{\tabcolsep}{1pt}
  \begin{tabular}{cccc}
    \includegraphics[width=0.245\linewidth]{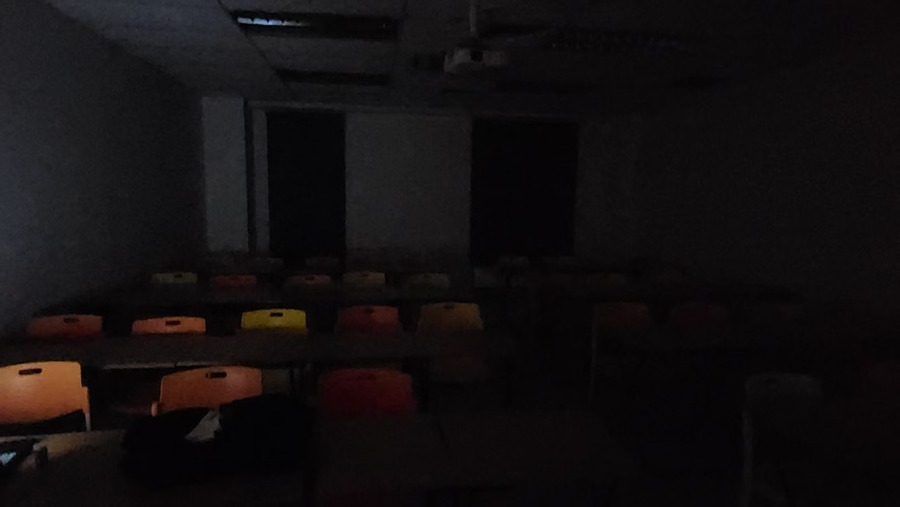} &
    \includegraphics[width=0.245\linewidth]{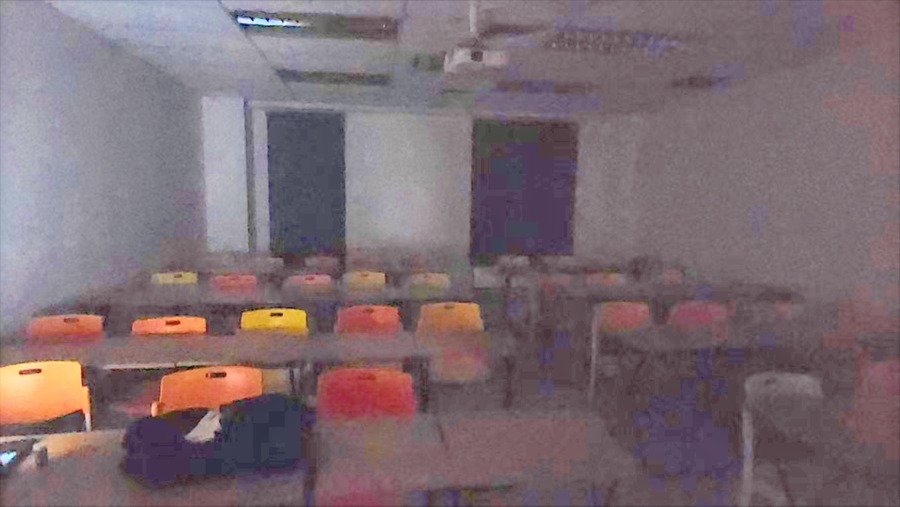} &
    \includegraphics[width=0.245\linewidth]{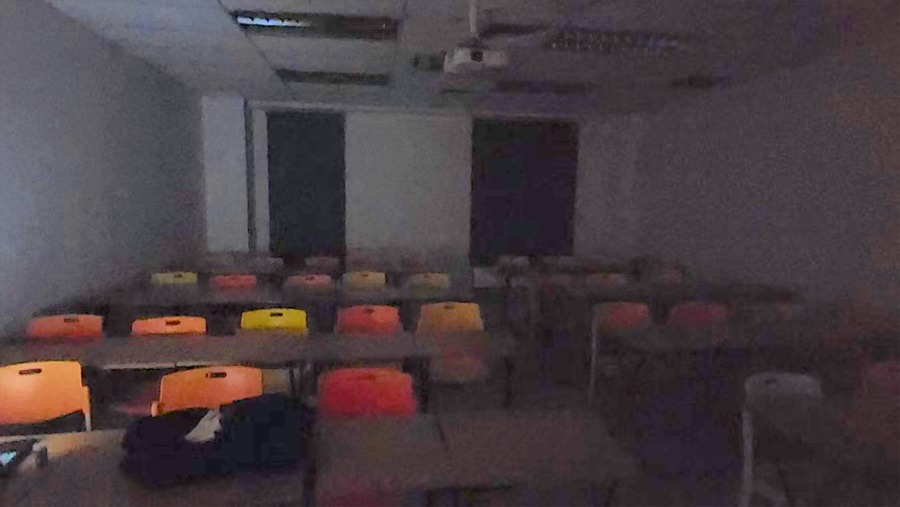} &
    \includegraphics[width=0.245\linewidth]{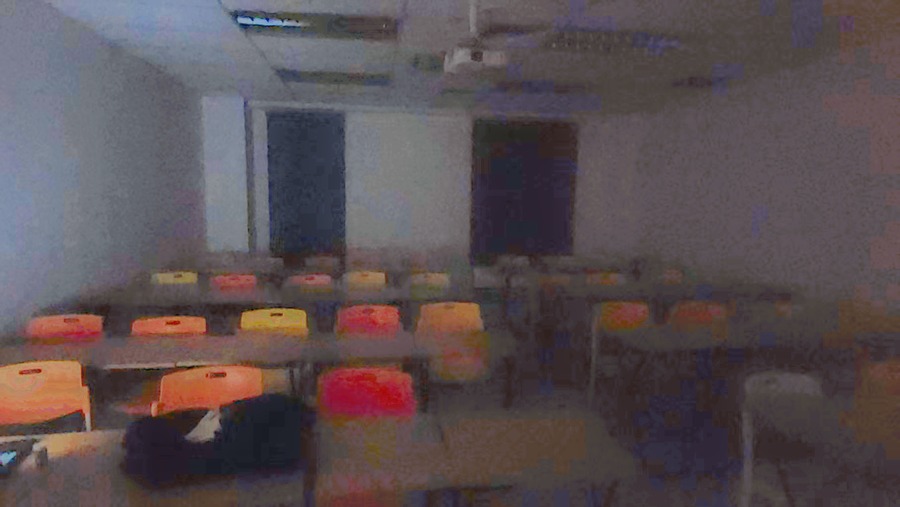} \\
    \includegraphics[width=0.245\linewidth]{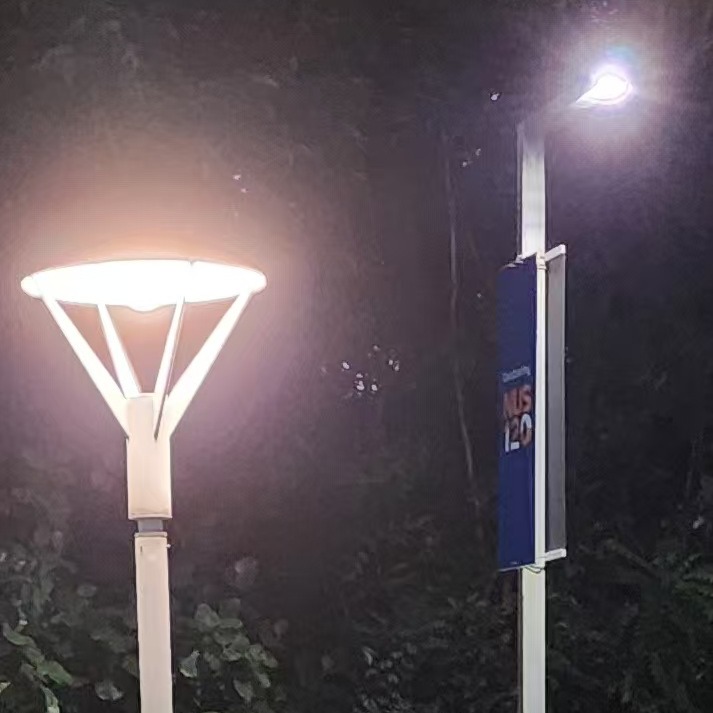} &
    \includegraphics[width=0.245\linewidth]{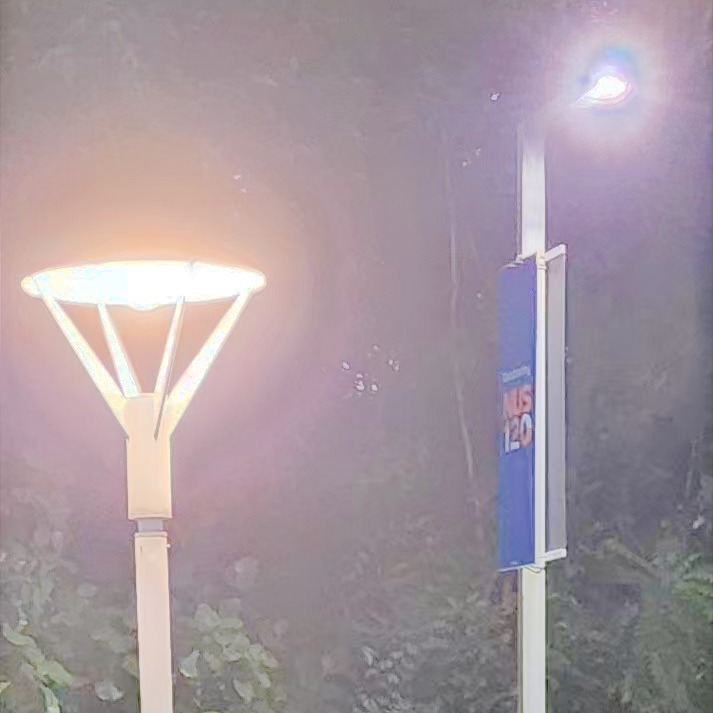} &
    \includegraphics[width=0.245\linewidth]{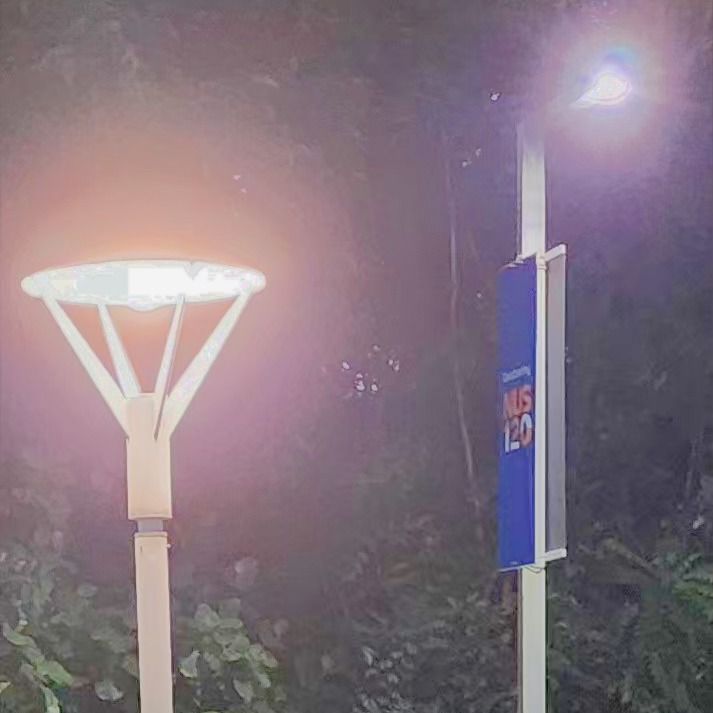} &
    \includegraphics[width=0.245\linewidth]{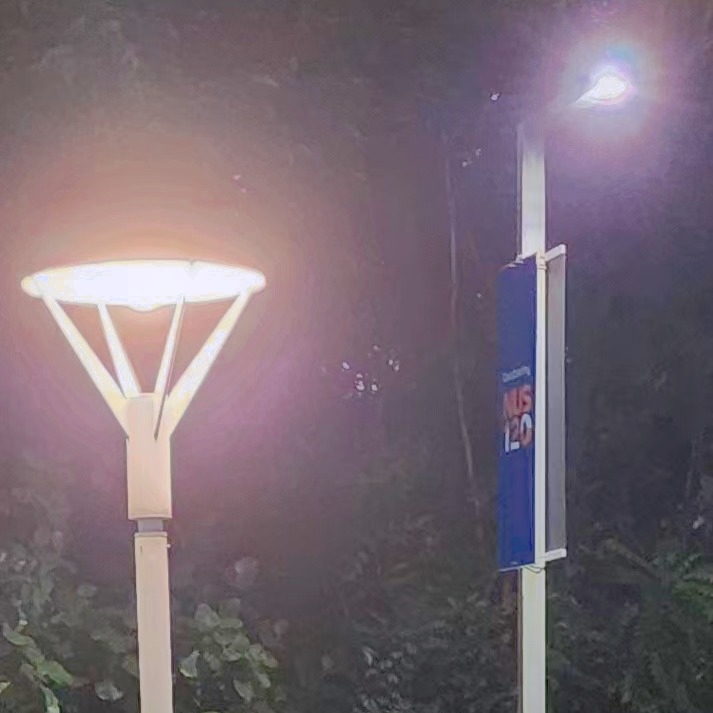} \\
    {\footnotesize Input} & {\footnotesize CSEC} & {\footnotesize LCDPNet} &
    {\footnotesize LoopLUT (ours)} \\
  \end{tabular}

  \vspace{6pt}
  \begin{minipage}[c]{0.44\linewidth}
    \centering
    \includegraphics[width=\linewidth]{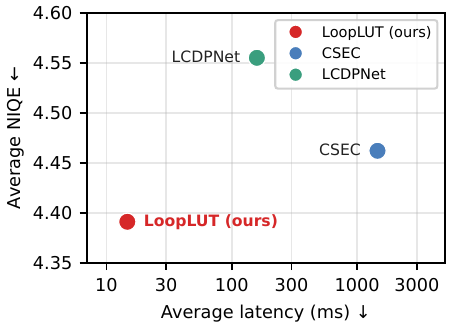}
  \end{minipage}\hfill
  \begin{minipage}[c]{0.54\linewidth}
    \centering
    \setlength{\tabcolsep}{4pt}
    \resizebox{\linewidth}{!}{%
    \begin{tabular}{@{}lccccc@{}}
      \toprule
      Method & Latency (ms)$\downarrow$ & Mem.\ (GB)$\downarrow$ & NIQE$\downarrow$ & BRISQUE$\downarrow$ & MUSIQ$\uparrow$ \\
      \midrule
      CSEC & 1455.9 & 3.71 & 4.46 & 33.27 & 34.67 \\
      LCDPNet & 158.8 & 3.62 & 4.56 & 33.57 & \textbf{37.35} \\
      LoopLUT (ours) & \textbf{14.7} & \textbf{0.12} & \textbf{4.39} & \textbf{32.74} & 37.18 \\
      \bottomrule
    \end{tabular}}
  \end{minipage}
  \caption{Real-world smartphone photographs. Top: an underexposed indoor scene;
  middle: a night scene with blown-out street lamps (same crop for all methods).
  Bottom: NIQE against latency (left) and averages over the two photographs
  (right; best in bold).}
  \label{fig:realworld}
\end{figure}

To examine performance in the real world beyond standard test sets, we took two
everyday photographs with a Samsung Galaxy S22 Ultra
(Figure~\ref{fig:realworld}): an underexposed indoor scene and a night scene
with blown-out street lamps, both ordinary JPEG files. The compared methods are
CSEC and LCDPNet, which are designed specifically for exposure correction.

LoopLUT improves the exposure of both scenes with fewer artifacts in highlight
regions such as the lamp shade, where CSEC produces cyan fringes and LCDPNet
produces blocky artifacts. Quantitatively, LoopLUT attains the best average
NIQE and BRISQUE, and its MUSIQ is only 0.17 below the best (LCDPNet).

On the same GPU and at the native input resolution, LoopLUT runs 11$\times$
faster than LCDPNet and 99$\times$ faster than CSEC, with about 1/30 of their
GPU memory. This shows that LoopLUT retains strong enhancement quality while
offering greater engineering value and suitability for edge devices.

\FloatBarrier
\section{Mobile deployment}
\label{sec:mobile}

\begin{figure}[!ht]
  \centering
  \includegraphics[width=0.95\linewidth]{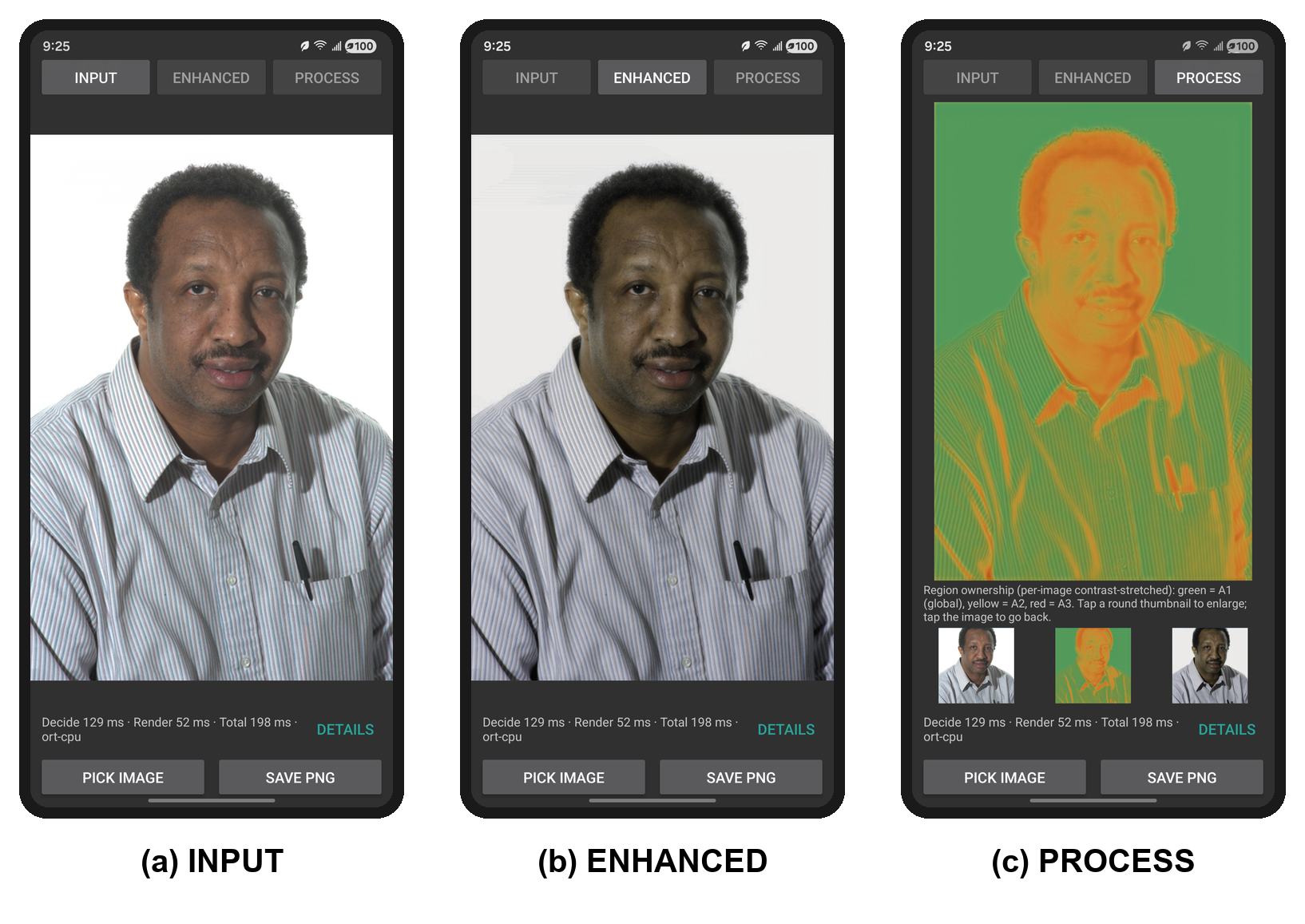}
  \caption{Mobile deployment. LoopLUT runs on an Android phone through the
  ONNX Runtime CPU backend, with three tabs corresponding to (a)~input,
  (b)~enhanced output and (c)~partition assignment. The colors in (c) are the
  partition weights of Equation~\eqref{eq:partition}: green for the global round
  $\mathbf{A}_1$, yellow for $\mathbf{A}_2$ and red for $\mathbf{A}_3$, displayed after per-image contrast
  stretching; the three thumbnails below can be tapped to enlarge. The bottom of
  the interface shows the measured readout for \textbf{that particular run}:
  Decide 129\,ms, Render 52\,ms, total 198\,ms. On-device total time fluctuates
  between roughly 100 and 200\,ms depending on the input photograph. Decide
  corresponds to the fixed-resolution decision stage and Render to the execution
  stage that grows with pixel count, matching the split in
  Table~\ref{tab:complexity}.}
  \label{fig:mobile}
\end{figure}

We export LoopLUT to ONNX and deploy it on an Android phone through the ONNX
Runtime CPU backend, giving a working enhancement application with three views:
input, enhanced output and partition assignment
(Figure~\ref{fig:mobile}). The partition view renders the partition of unity of
Equation~\eqref{eq:partition} on the device, making the per-round contraction
visible on real hardware rather than only offline.

On \textbf{CPU alone} the total time per frame fluctuates between roughly 100
and 200\,ms depending on the photograph. This is \textbf{not directly
comparable} with the 5.10\,ms result of
Table~\ref{tab:complexity}, since the hardware and backend differ entirely and
no GPU or NPU acceleration is used. The on-device split does reproduce the core
property of the architecture: Decide and Render are measured separately, and
since the decision stage runs only at a fixed $256\times 256$, raising the
output resolution increases only Render. The model is exported in fp32 with
11.04\,MB of weights, without quantization or pruning, so these are the same
weights as in Table~\ref{tab:main}.

\end{document}